\documentclass{article}

\usepackage{arxiv}
\usepackage[utf8]{inputenc} 
\usepackage[T1]{fontenc}    
\usepackage{hyperref}       
\usepackage{url}            
\usepackage{booktabs}       
\usepackage{amsfonts}       
\usepackage{nicefrac}       
\usepackage{microtype}      

\usepackage{algorithm}
\usepackage{algorithmic}
\usepackage{graphicx}
\usepackage{subfig}

\usepackage{xcolor}         
\usepackage{amsmath}
\usepackage{fourier}
\usepackage{float}

\DeclareMathAlphabet{\mathcal}{OMS}{cmsy}{m}{n}
\SetMathAlphabet{\mathcal}{bold}{OMS}{cmsy}{b}{n}

\title{Single-Shot Pruning for Offline Reinforcement Learning}

\author{
  Samin Yeasar Arnob \\
  Mila, McGill University\\
  \texttt{samin.arnob@mail.mcgill.ca} \\
   \And
   Riyasat Ohib \\
   Georgia Institute of Technology \\
   Atlanta, GA \\
   \texttt{riyasat.ohib@gatech.edu} \\
   \AND
   Sergey Plis \\
   Georgia State University \\
    Atlanta, GA \\
   \texttt{s.m.plis@gmail.com} \\
   \And
   Doina Precup \\
   Mila, McGill University,\\
   DeepMind\\
}

\begin{document}
\maketitle

\begin{abstract}
  Deep Reinforcement Learning (RL) is a powerful framework for solving complex real-world problems.
  Large neural networks employed in the framework are traditionally associated with better generalization capabilities, but their increased size entails the drawbacks of extensive training duration, substantial hardware resources, and longer inference times. One way to tackle this problem is to prune neural networks leaving only the necessary parameters.
  State-of-the-art concurrent pruning techniques for imposing sparsity perform demonstrably well in applications where data-distributions are fixed.
  However, they have not yet been substantially explored in the context of RL.
  We close the gap between RL and single-shot pruning techniques and present a general pruning approach to the Offline RL. We leverage a fixed dataset to prune neural networks before the start of RL training.
  We then run experiments varying the network sparsity level and evaluating the validity of \textit{pruning at initialization} techniques in continuous control tasks. Our results show that with $95\%$ of the network weights pruned, Offline-RL algorithms can still retain performance in the majority of our experiments.
  To the best of our knowledge no prior work utilizing pruning in RL retained performance at such high levels of sparsity.
  Moreover, \textit{pruning at initialization} techniques can be easily integrated into any existing Offline-RL algorithms without changing the learning objective.

\end{abstract}

\keywords{OfflineRL \and Sparse Network \and Pruning}

\section{Introduction}


Function approximation with deep neural networks has been extremely successful in the last decade for a range of complex tasks. However, this performance comes at a significant computational cost and excessive memory requirements due to their good optimization and generalization performance in the highly overparamaterized region \cite{zhang2021understanding, neyshabur2018role, arora2019fine, zhang2019fast}. It is an attractive proposition to prune such large networks with negligible loss in performance for real-time applications specially on edge devices with resource constraints. Pruning of networks \cite{lecun1990optimal, hassibi1993optimal, dong2017learning, han2015learning} has demonstrated reduction in inference-time resource requirements with minimal performance loss. The standard approach is to prune network after training and then perform a costly retraining step, thus requiring an extended training regime to generate sparse networks. Moreover, it is also difficult to train sparse networks from scratch which maintains similar performance to their dense counterparts \cite{han2015deep, li2016pruning}. Although, pruning before training is difficult, there are significant benefits in time and resource efficiency if we can prune networks before training. 

In recent times, the Lottery Ticket Hypothesis \cite{frankle2018lottery} was proposed that details the presence of sub-architectures within a larger network, which when trained are capable of reaching the baseline accuracy of the dense networks or even surpass them in some cases. The sparsity in these sub-architectures can be exploited with suitable hardwares for computational efficiency gains, such as in \cite{dey2019pre}, where authors managed to demonstrate a $5\text{x}$ efficiency gain while training networks with pre-specified sparsity.


There are a range of techniques in literature that provide methods to prune Deep Neural Networks at various stages of their training and testing. The most common of these techniques is to prune the network after training using some sort of predefined criterion that captures the significance of the parameters of the network to the objective function. A range of classical works on pruning used the second derivative information of the loss function \cite{lecun1990optimal, hassibi1993optimal}. Perhaps the most intuitive of these approaches is magnitude pruning, where following training a subset of the parameters below some threshold is pruned and the rest of the parameters are retrained \cite{han2015deep, han2015learning} and regularization based methods \cite{yang2019deephoyer, ma2019transformed, louizos2017learning, yun2019trimming} which induces sparsity in the network during the optimization process.

Other more elaborate techniques to find the lottery tickets include solving a separate optimization problem to find out the subset of weights of the lottery-ticket the sub-architecture \cite{zhang2018systematic, gillis2019grouped, li2019compressing}. However, pruning of randomly initialized network before training still seemed like a difficult task, as the connections of a randomly initialized network exhibits little information about their significance to the training process. This, however, was changed with the proposal of \emph{Single-shot Network Pruning} (SNIP) \cite{lee2018snip}, which managed to prune weights before training with great success by finding out sparse trainable sub-architectures. With SNIP it has been demonstrated that it is indeed possible to prune neural networks in one-shot at initialization. A recent work \cite{wang2020picking} challenges SNIP's pruning criterion of \textit{connection sensitivity} and argues that this is sub-optimal as the gradient of each weight is susceptible to change after pruning due to complex interactions among weights. Therefore, with this technique there is a possibility of pruning weights that are vital for the flow of information through the network. Instead, the authors of \cite{wang2020picking} propose an alternative method, \emph{Gradient Signal Preservation} (GraSP), that preserves the gradient flow of the network. These techniques where the network weights are pruned before training can be termed as \textit{pruning at initialization}. There are a few recent works \cite{DST-RL, Pops} that leverages different pruning techniques in Deep-RL algorithms but they prune the neural-network in-between Online-RL training. Since the RL agent gets updated in every iteration and collects data through environment interactions, there are significant shifts in the data-distribution. Thus, it makes harder for the pruning techniques to find the proper sub-networks that can perform the same. To the best of our knowledge, current state-of-the-art pruning in Online-RL methods can sparsify the networks up to $50\%$ without sacrificing performance \cite{DST-RL}. But in this work, we show we can do better.

Similar to supervised training, Offline-RL is trained with fixed dataset. Therefore, pruning techniques that are suitable for fixed dataset can be used in offline-RL as well. In this work, we explore single-shot pruning techniques in Offline-RL algorithms. This allow us to prune the networks even before we start training an RL agent. Up until very recently batch-dataset RL in the setting of continuous control was presumed to be a hard problem. This is due to not having access to environment interactions and RL agents needing to learn from a fixed dataset. But, we can instead leverage this fixed-dataset nature of offline-RL to apply one-shot pruning techniques that are not suitable for online-RL algorithms.

Through this work we want to excite the community more about offline-RL research and pruning techniques in RL algorithms. Our contributions in this work are as follows:
\raggedbottom
\begin{itemize}
    \item In this work, we show experimental results of pruning methods in Offline-RL algorithms where we use the following one-shot pruning methods: \emph{SNIP} \cite{lee2018snip} and \emph{GraSP} \cite{wang2020picking}. We explain how these single-shot pruning methods can be integrated with Offline-RL algorithms.
    \item  We demonstrate it is possible to prune $95 \%$ of the network parameters without losing performance in continuous control tasks. 
    \item We also show that it is possible to reduce the memory required to store these pruned networks by $4\text{x}$ without any elaborate compression mechanism.
\end{itemize}

\section{Preliminaries}

We consider learning in a Markov decision process (MDP) described by the tuple ($S, A, P, R$). The MDP tuple consists of states $s \in S$, actions $a\in A$, transition dynamics $P(s'|s,a)$, and reward function $r=R(s,a)$. We use $s_t$, $a_t$ and $r_t=R(s_t,a_t)$ to denote the state, action and reward at timestep t, respectively. A trajectory is made up of sequence of states, action and rewards $\tau=(s_0, a_0, r_0, s_1, a_1, r_1, ..., s_T,a_T,r_T)$. For continuous control task we consider an infinite horizon, where $T=\infty$ and the goal in reinforcement learning is to learn a policy which maximizes the discounted expected return $\mathbb{E}[\sum_{t=t'}^T \gamma^t r_t]$ in an MDP. In offline reinforcement learning, instead of obtaining data through environment interactions, we only have access to some fixed limited dataset consisting of trajectory rollouts of arbitrary policies. This setting is harder for agent as it can not further explore the environment and collect additional feedback. Thus can fail due to overestimation
of values induced by the distributional shift between the dataset and the learned policy. Offline algorithms \cite{BCQ, TD3_BC, CQL, fisher_RBC} overcome the problem through either constraining policy or the value function estimation. 


\section{Methods of pruning at initialization}
In this section, we discuss two methods of pruning at initialization namely, SNIP and GraSP and briefly discuss their criterions of pruning. A more elaborate discussion is available at \cite{lee2018snip, wang2020picking}.

\subsection{Single-shot Network Pruning at initialization}
The first work to tackle pruning at initialization was SNIP \cite{lee2018snip} which exploits the idea of \textit{connection sensitivity} to prune insignificant weights. They formalize this idea in terms of removing a single weight $\theta_q$ and the effect it has on the loss as:

\begin{equation}
    S(\theta_q) = \lim_{ \epsilon \to 0} \left| \frac{\mathcal{L}(\theta_0) -\mathcal{L} (\theta_{0} + \epsilon \delta_q)}{\epsilon} \right| = \left| \theta_q \frac{\partial \mathcal{L}}{\partial \theta_q} \right|
\end{equation}

where $\theta_q$ corresponds to the $q^{th}$ element of $\theta_0$, and $\delta_q$ is a one-hot vector whose $q_{th}$ element equals to $\theta_q$. The goal of SNIP is to essentially preserve the loss of the randomly initialized network before training. Although the idea to preserve the loss value was behind some classic works in pruning \cite{lecun1990optimal, hassibi1993optimal}, its importance is less obvious for pruning at initialization before the training begins. The authors of GraSP \cite{wang2020picking} instead argue that it is more important to preserve the training dynamics during pruning before training rather than the loss itself, because with the first technique there is a chance to make some layers too sparse that creates a bottleneck in the neural network for signal propagation. Therefore, they argue that a pruning technique i.e. Gradient Signal Preservation, that takes into account how the presence of a connection affects the training of the whole network would be preferable. 

\subsection{Gradient Signal Preservation}
The idea of utilizing \emph{Gradient Signal Preservation} (GraSP) to improve upon the work of SNIP was presented in the work \cite{wang2020picking} with the algorithm the authors termed as GraSP. Pruning a network results in fewer parameters and reduced connectivity which might lead to a decrease in the flow of gradients through the network thus slowing down the optimization process. More formally, a larger norm of the gradient points to each gradient update contributing towards a greater loss reduction to the first order, as indicated by the directional derivative:

\begin{align} \label{eq:2}
    \Delta \mathcal{L} (\theta) = 
    \lim_{ \epsilon \to 0} \frac{\mathcal{L} (\theta + \epsilon \nabla \mathcal{L} (\theta)) - \mathcal{L} (\theta)}{\epsilon}
    = \nabla \mathcal{L} (\theta)^T \nabla \mathcal{L} (\theta)
\end{align}

The goal of GraSP is to preserve (even increase if possible) the gradient flow after pruning the network. Similar to the classic work \cite{lecun1990optimal} the authors cast the pruning operation as adding a perturbation $\delta$ to the initial weights. A Taylor approximation is then used to characterize the effect of removing one weight to the gradient flow through the network. 

\begin{align} \label{eq:3}
    \mathbf{S}(\delta)  
   =& \Delta \mathcal{L} (\theta_0 + \delta) - \Delta \mathcal{L} (\theta_0) \nonumber \\
   =& 2 \delta^T \nabla^2 \mathcal{L} (\theta_0) \nabla \mathcal{L} (\theta_0 ) + \mathcal{O}( \Vert \delta \Vert_2^2) \nonumber \\ =& 2 \delta^T \mathbf{Hg} + \mathcal{O}( \Vert \delta \Vert_2^2)
\end{align}

where $ \mathbf{S}(\delta)$ is an approximate measure of the change to equation \ref{eq:2}. The dependencies among the parameters of the network is captured by the Hessian matrix, which acts as a predictor of the effects of removing multiple weights. 

GraSP essentially uses equation \ref{eq:3} to calculate the score of each weight corresponding to its effect on the reduction of gradient flow after pruning. More precisely, a negative $S(\delta)$ will correspond to a reduction of gradient flow if the associated weight is pruned, while a positive value will result in an increase of gradient flow if said weights are pruned. Therefore the larger the scores associated with the weights, the lower their importance and those weights are removed first. Therefore the vectorized scores are calculated as:

\begin{align}
   \mathbf{S} (- \mathbf{\theta}) = -\mathbf{\theta} \odot \mathbf{Hg}
\end{align}

GraSP then removes the \textit{top k} fraction of the weights for a given pruning ration of \textit{k} to generate a pruning mask by computing the scores associated with each weight. Thus, GraSP takes the gradient flow into account while pruning the network.

\begin{algorithm}[h!]
\caption{Single-shot Pruned Offline-RL Training}
  \label{algorithm_1}
 \begin{algorithmic}
     \STATE  
     \textbf{Initialize Networks}: critic $Q_{\theta_1}$, $Q_{\theta_2}$, Actor $\pi_\phi$, VAE $V_{\omega}=\{V_{\omega_E}, V_{\omega_D}\}$ \\ 
     \textbf{Choose single shot pruning technique}: SNIP or GraSP \\
     
     \textbf{Find the Pruning Weight Maps}: \STATE $M_{\theta_1}$, $M_{\phi}$, $M_{\omega_E}$, $M_{\omega_D}$  = \text{single-shot pruning} $\Big( \mathcal{L}(\theta_1) , \mathcal{L}(\phi), \mathcal{L}(\omega_E), \mathcal{L}(\omega_D)\Big)$ \\
     
     \textbf{Prune the networks:} \\
     $\theta_1 \leftarrow \theta_1 \odot M_{\theta_1} $, \\ 
     $\theta_2 \leftarrow \text{copy}(\theta_1) $, \\
     $\phi \leftarrow \phi \odot M_{\phi} $, \\
     $\omega_E \leftarrow \omega_E \odot M_{\omega_E} $, \\
     $\omega_D \leftarrow \omega_D \odot M_{\omega_D}$  \\ 
  \FOR{$t=1$ {\bfseries to} $T$}
     \STATE Train Offline-RL algorithm
     \ENDFOR
 \end{algorithmic}
\end{algorithm}
\section{Experiments}

We perform our experiments on OpenAI Gym MuJoCo continuous control tasks \cite{mujoco, OpenAi_gym} on two different Offline-RL algorithms: \emph{Batch-Constrained
deep Q-learning} (BCQ) \cite{BCQ} and \emph{Behavior Cloning} (BC) (implemented in \cite{Bear}). Without changing anything within the RL objective, we integrate following two pruning approaches: \emph{SNIP} \cite{lee2018snip} and \emph{GraSP} \cite{wang2020picking}. These one-shot pruning methods find the important neural-network weights before initializing RL training loop and set the rest of the weights to zero which remains zero through-out the RL training. For example, to train a $95\%$ sparse network, we will set the $95\%$ weights to zero using one of these pruning techniques and will train the RL agent with the remaining $5\%$ of the weights which the pruning methods finds to be more relevant to the RL learning objective. 

We run our experiment varying different sparsity levels to understand at what extend we can reduce the model without sacrificing the performance. Since the one-shot pruning methods are independent of the Offline-RL objective, we expect these pruning methods to perform the same for other Offline-RL algorithms as well. 


BCQ and BC share the same architecture, where they use separate Actor, Critic and a VAE neural network. Before starting the network training, we sample single batch of training samples (100 random $(s,a,s',r)$ for SNIP and 200 random $(s,a,s',r)$ samples for GraSP) to generate a pruning mask. We then prune the neural network with these mask and train the remaining weights. The complete process is detailed in Algorithm-\ref{algorithm_1}.  Actor, Critic and VAE network have different objective functions. We individually compute these maximization/minimization objectives to find out the weights that are most relevant in optimizing these objectives.

\begin{figure}[hbt!]
\centering
   \includegraphics[width=0.32\linewidth]{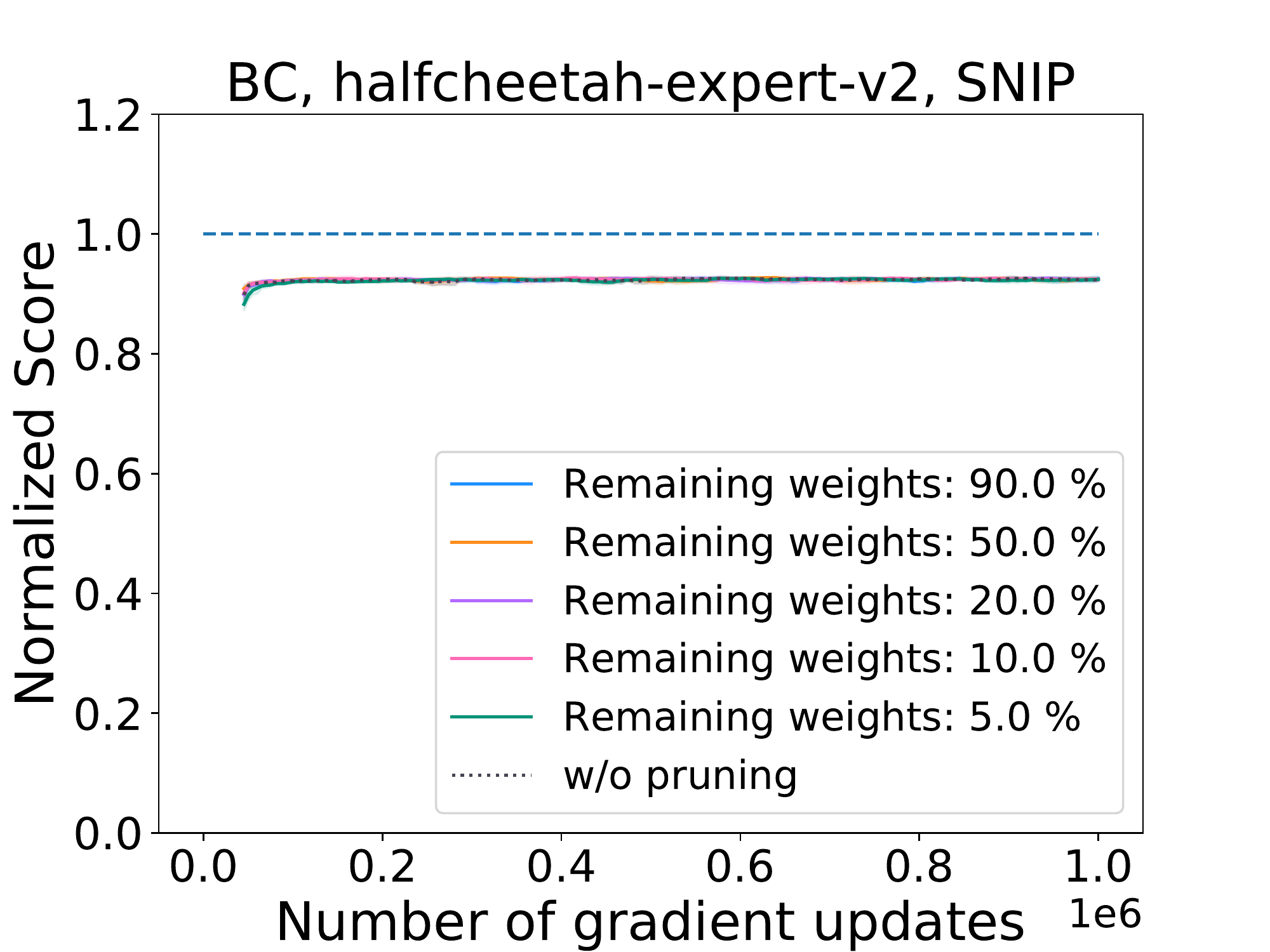}
   \
    \includegraphics[width=0.32\linewidth]{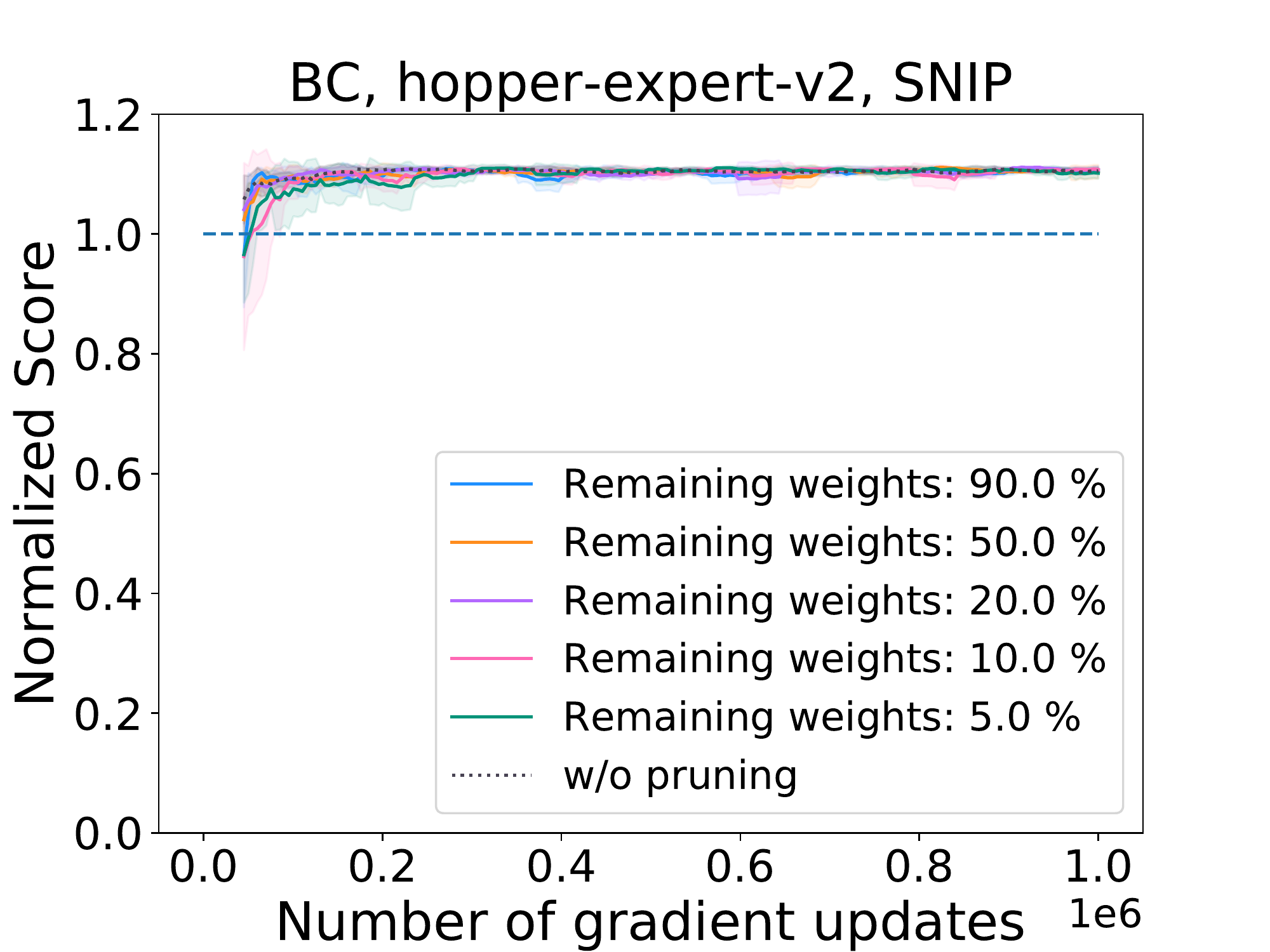}
   \
    \includegraphics[width=0.32\linewidth]{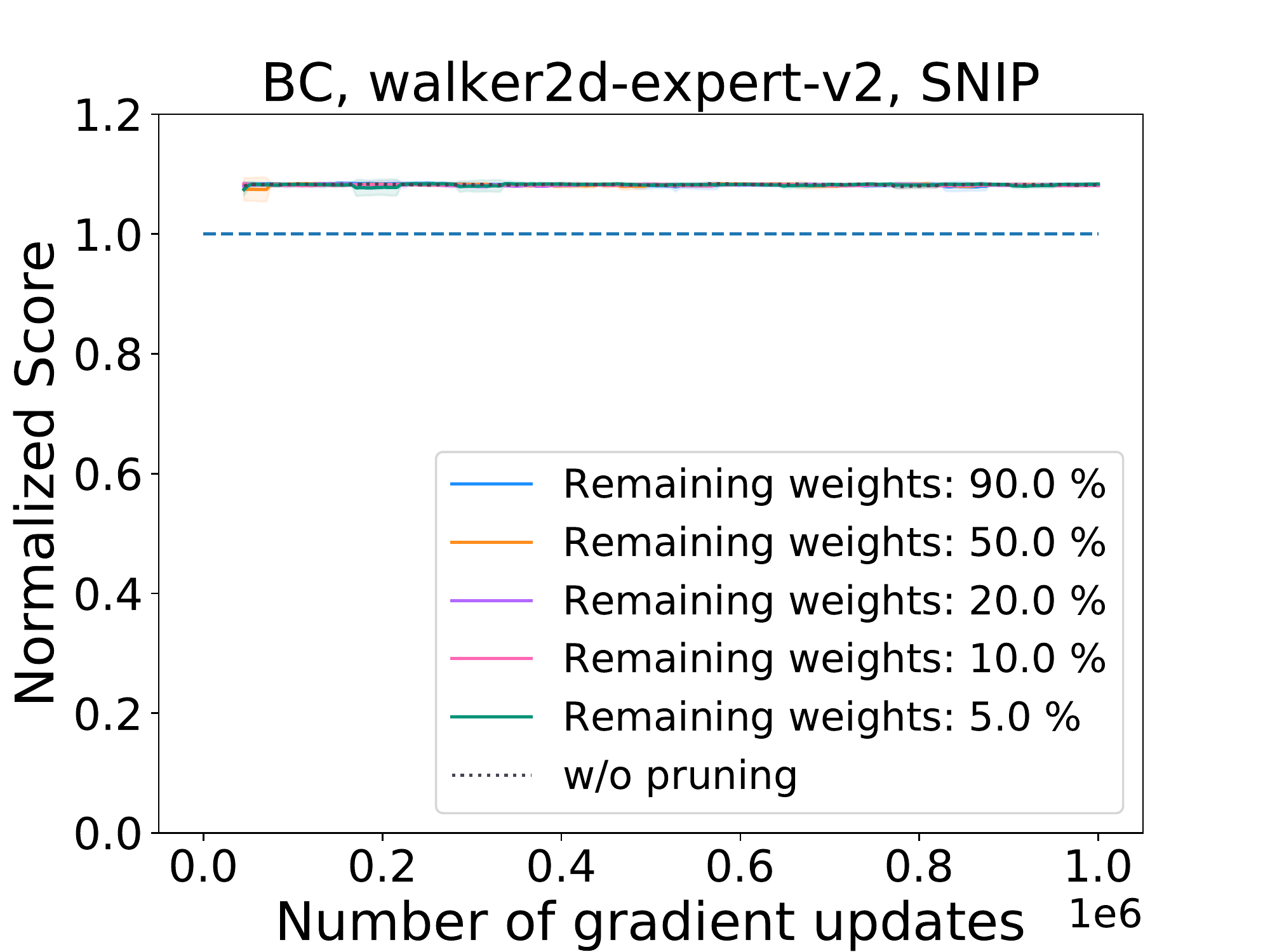}
    \
    
     \includegraphics[width=0.32\linewidth]{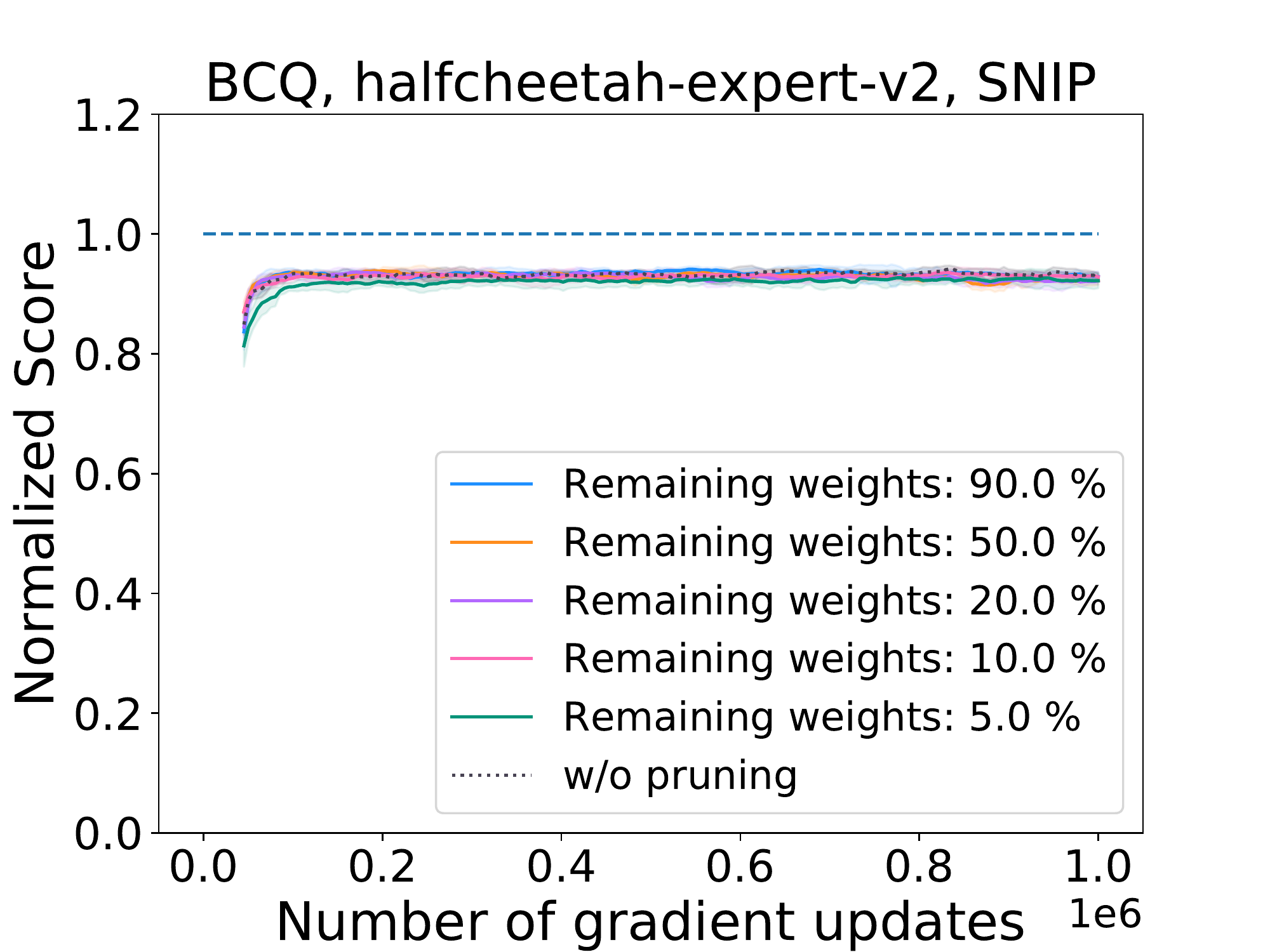}
   \
    \includegraphics[width=0.32\linewidth]{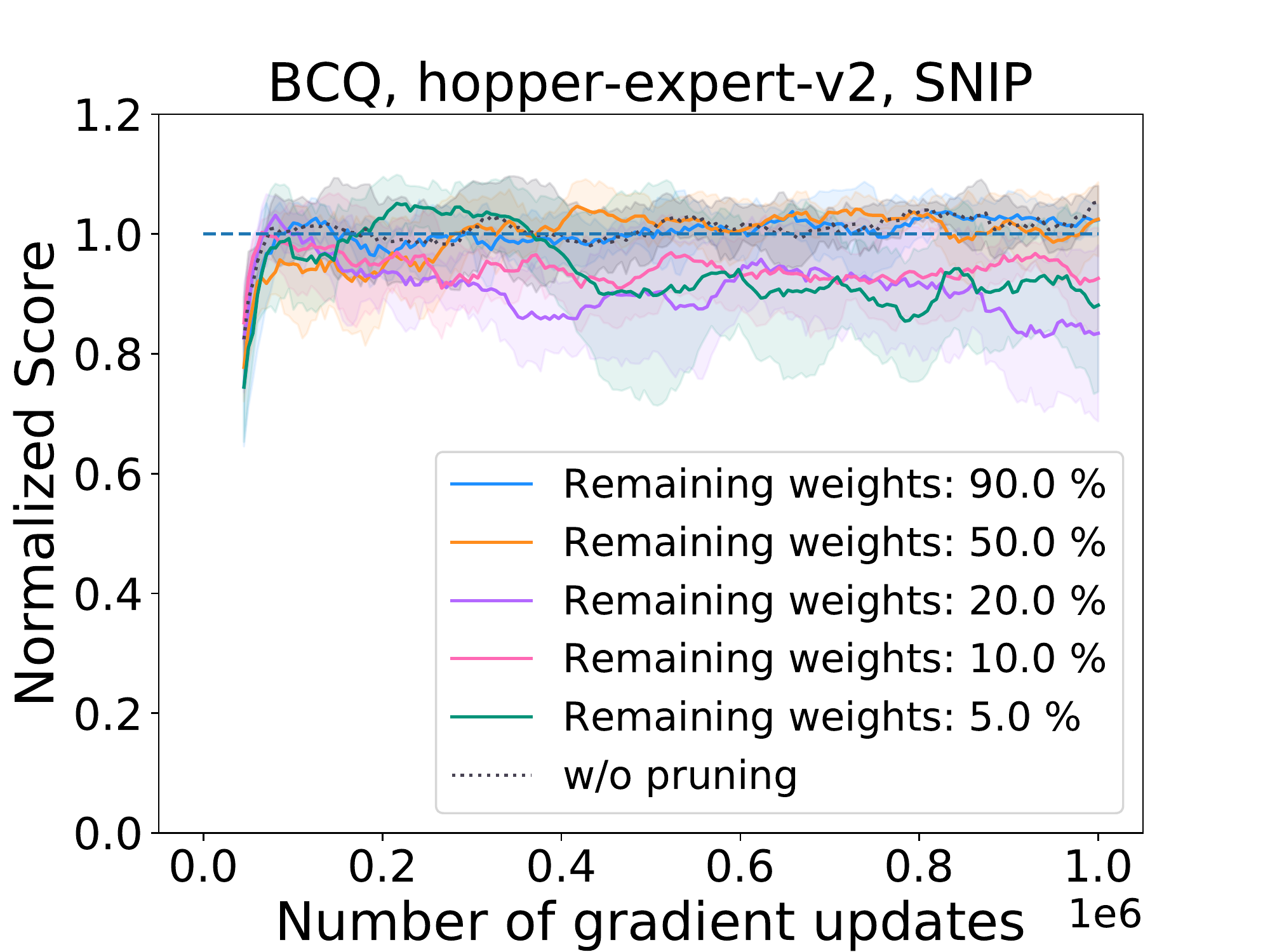}
   \
    \includegraphics[width=0.32\linewidth]{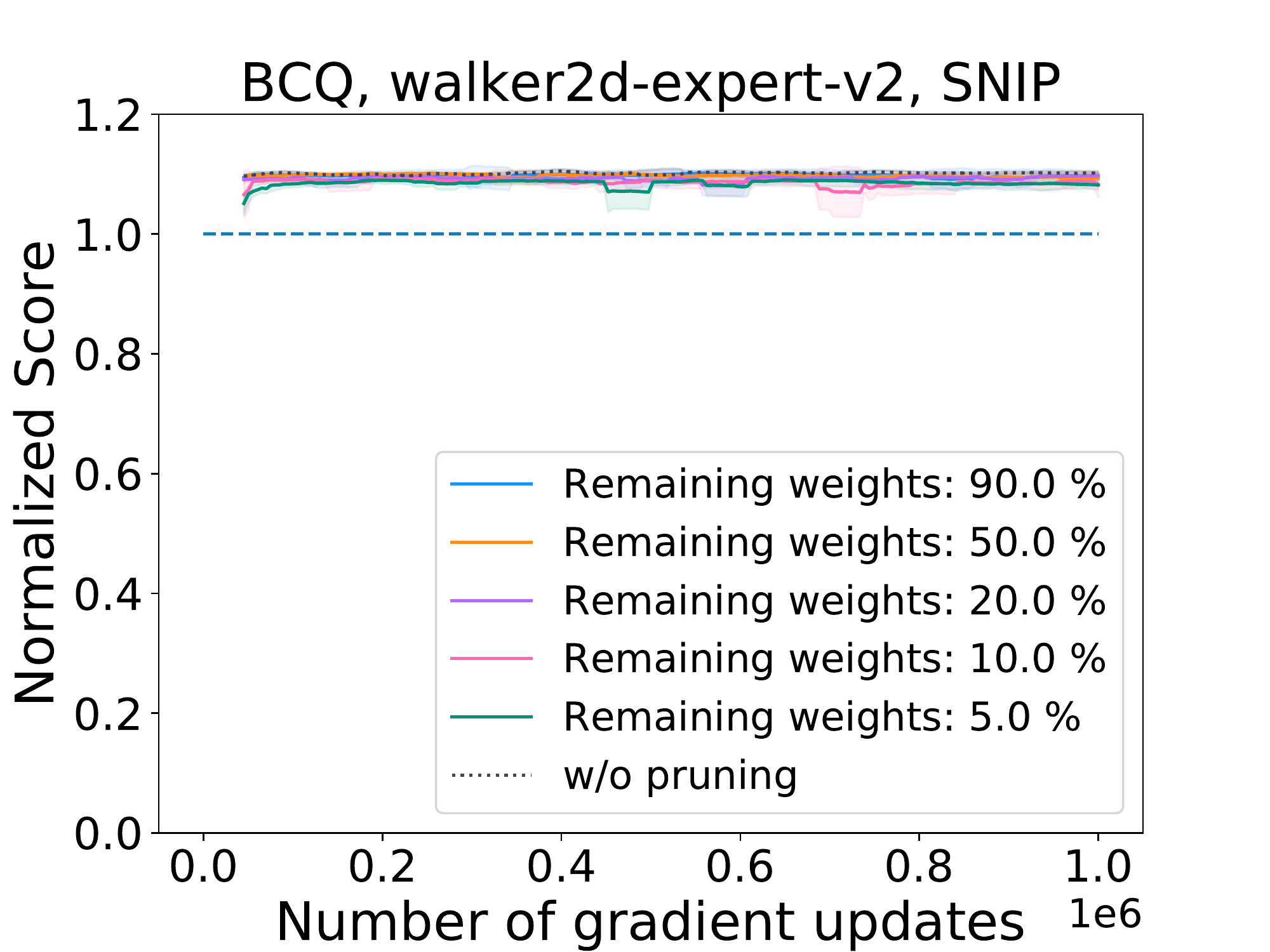}
\caption{Performance plot of Offline-RL algorithms (BCQ, BC) varying sparsity using SNIP}
\label{SNIP vary sparsity}
\end{figure}

\begin{figure}[hbt!]
\centering
   \includegraphics[width=0.32\linewidth]{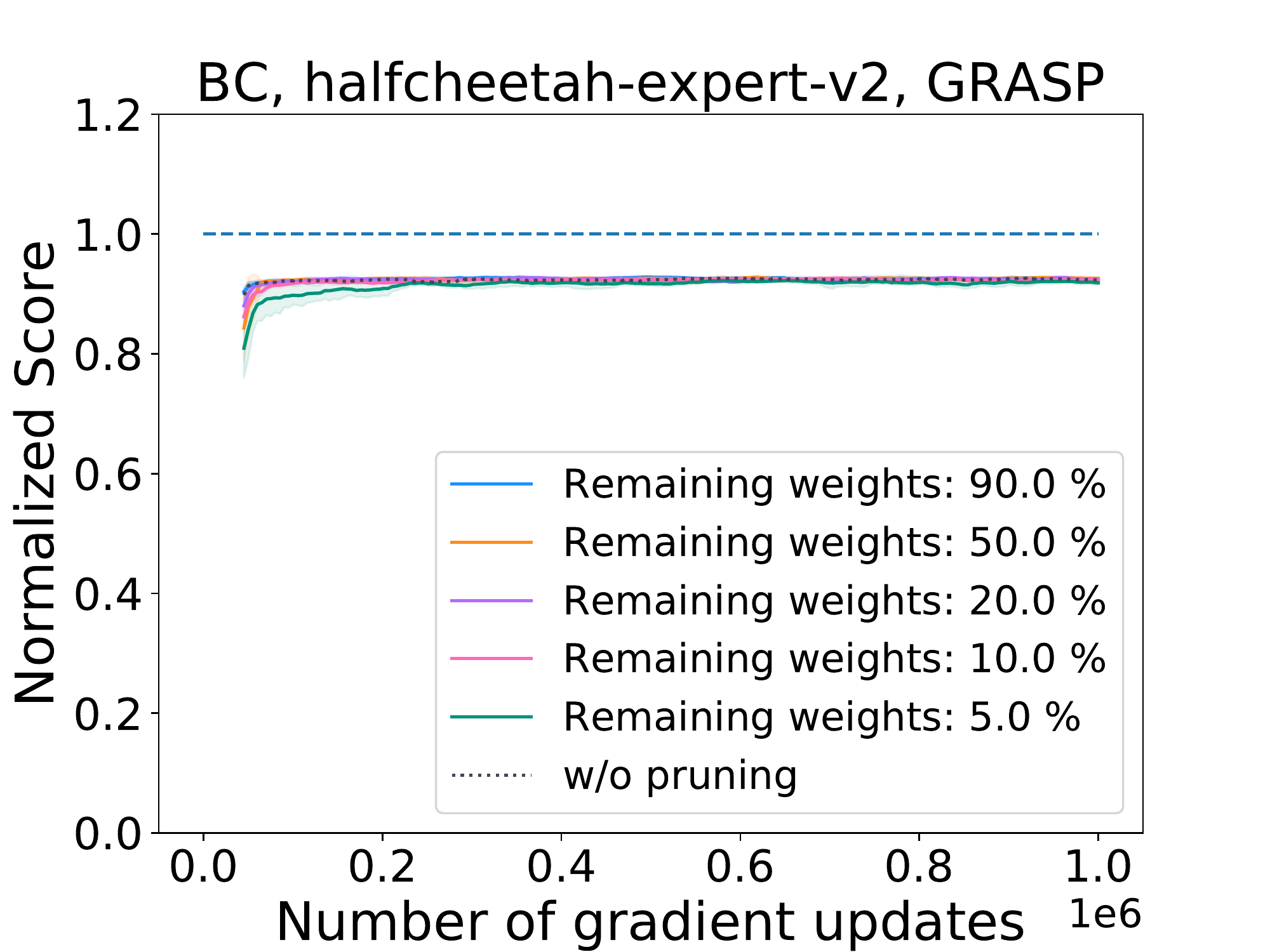}
   \
    \includegraphics[width=0.32\linewidth]{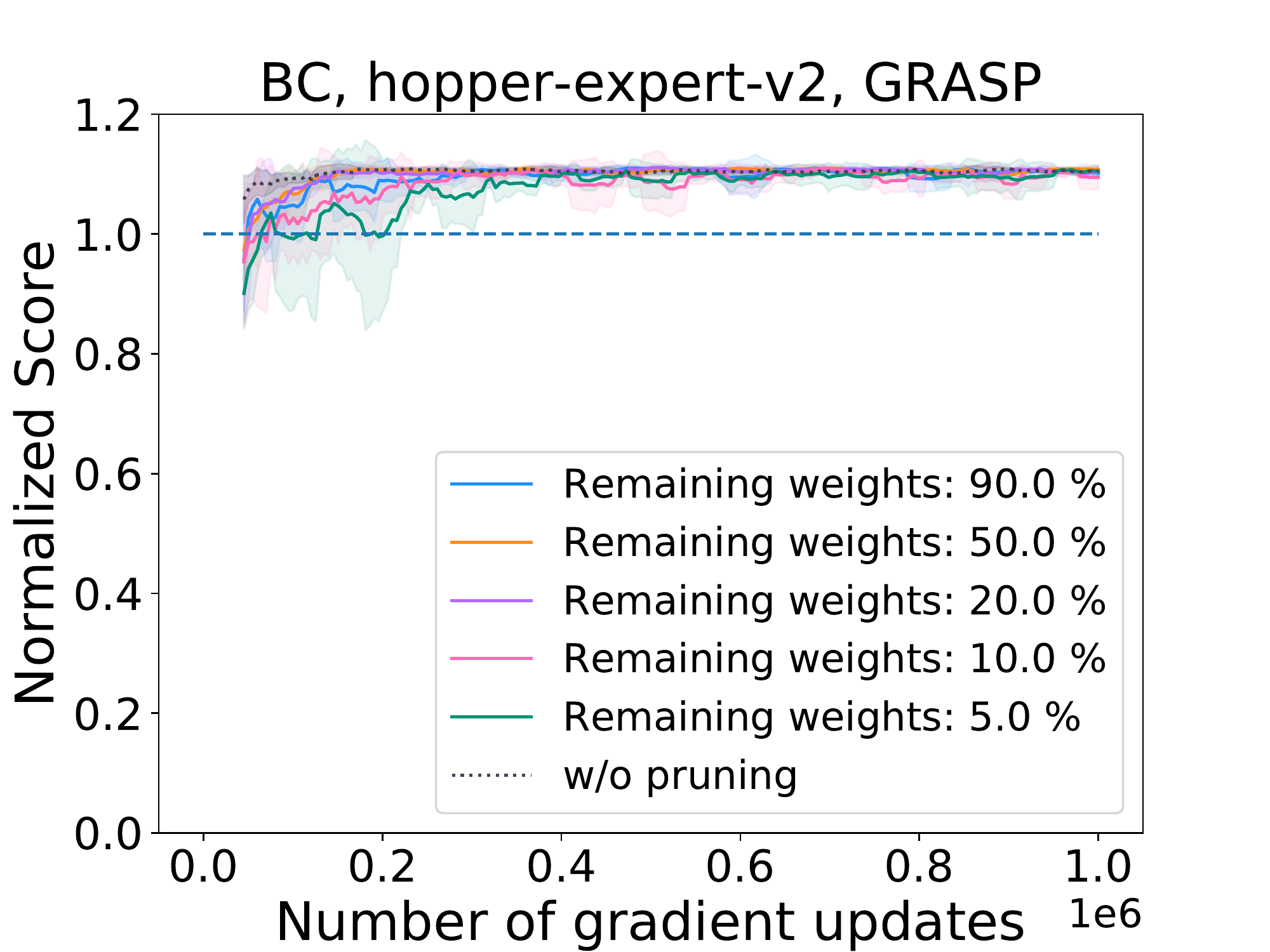}
   \
    \includegraphics[width=0.32\linewidth]{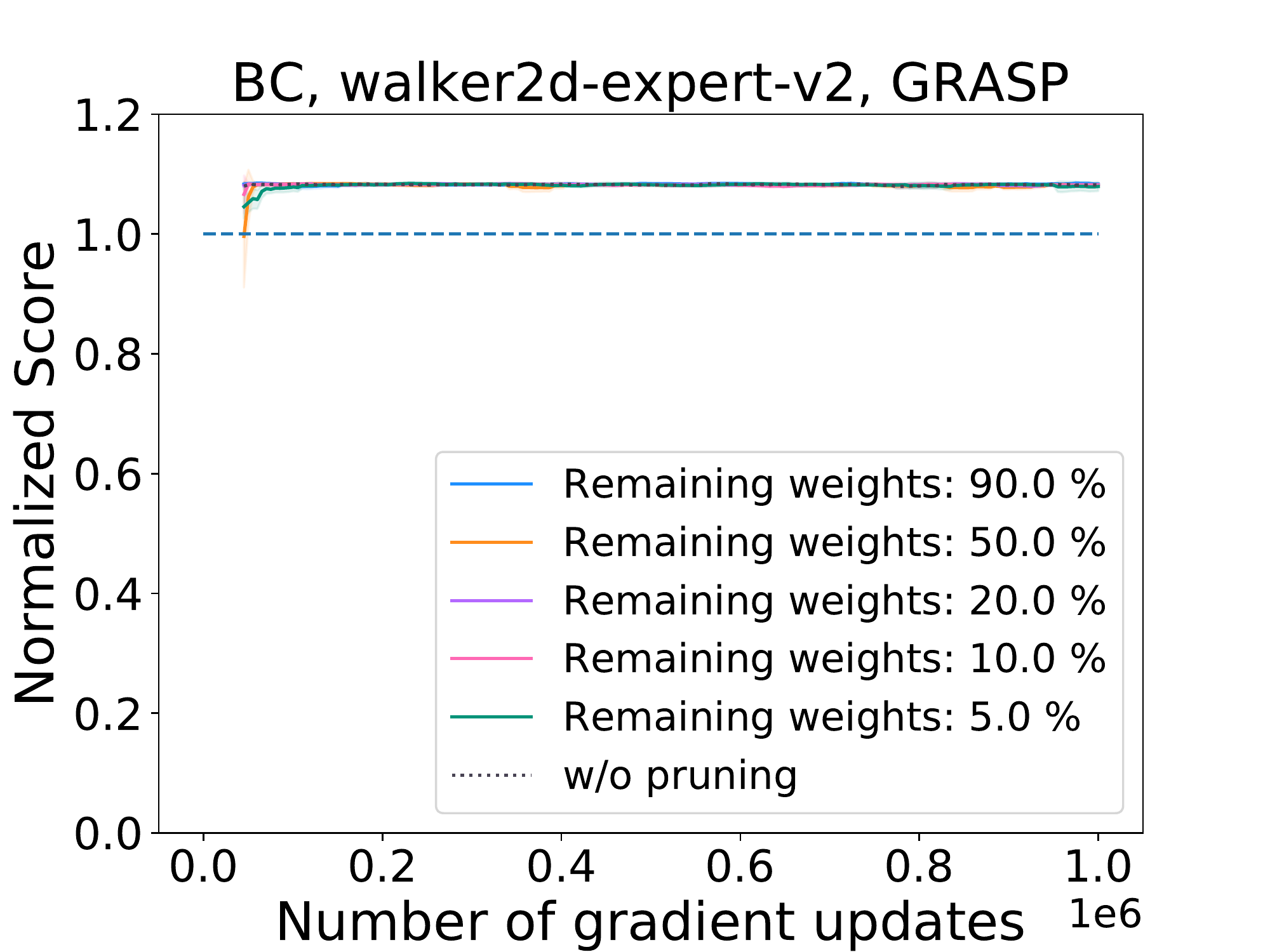}
    \
    
     \includegraphics[width=0.32\linewidth]{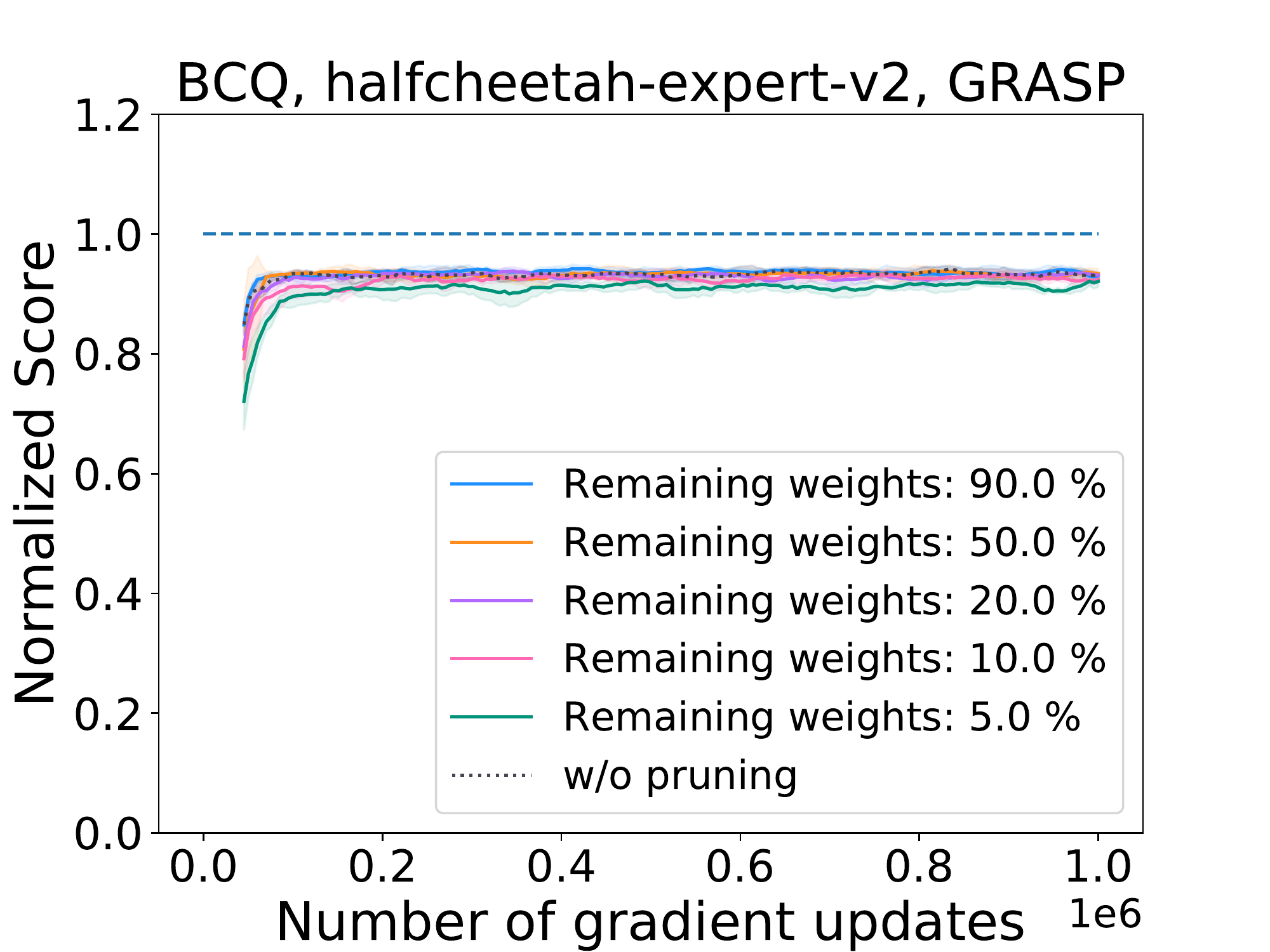}
   \
    \includegraphics[width=0.32\linewidth]{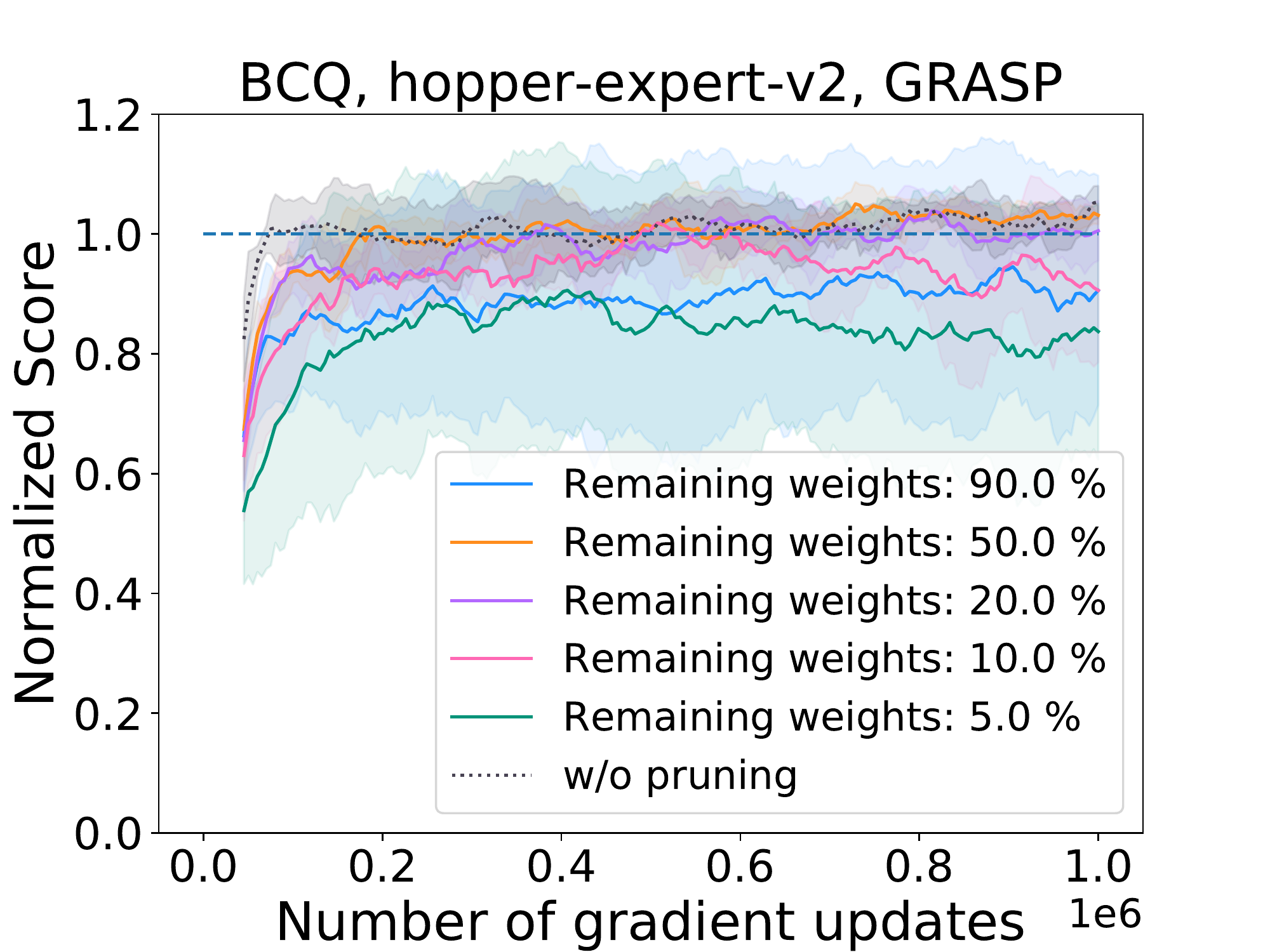}
   \
    \includegraphics[width=0.32\linewidth]{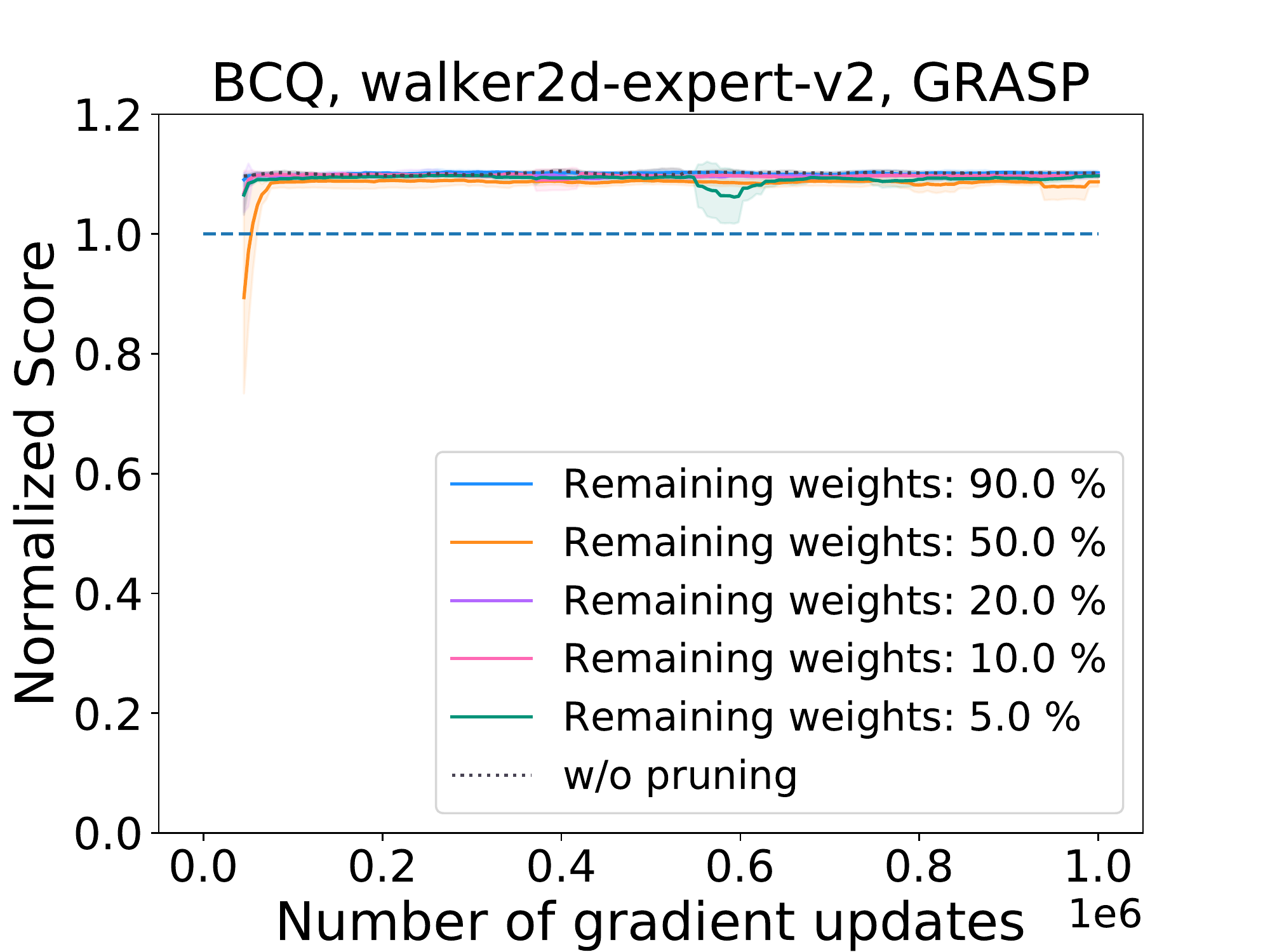}
\caption{Performance plot of Offline-RL algorithms (BCQ, BC) varying sparsity using GraSP}
\label{GRASP vary sparsity}
\end{figure}

We vary the sparsity of these networks from $10\%$ to $95\%$ and compare the performance of the Offline-RL algorithms. We perform our experiments on \emph{Half-Cheetah-v2}, \emph{Hopper-v2}, \emph{Walker2d-v2} environments and train the offline-RL algorithms with \emph{D4RL} \cite{D4RL} expert dataset. We plot the mean performance for seeds $0-4$ over 1 million gradient updates with $100\%$ confidence interval. In figures \ref{SNIP vary sparsity} and \ref{GRASP vary sparsity} we observe that, for all the experiments, except for BCQ in Hopper task, the performance in consistent even with $95\%$ sparse networks. This means with a fraction number of weights we still attain the performance of a large neural network. The pruning techniques can find the "sub-networks" that gives similar performance using only $5\%$ parameters of the larger network.

This is important to note that \emph{Dynamic Sparse TD3} (DS$-$TD3)  \cite{DST-RL}, a very recent research work hat uses sparse training with dynamic sparsity \cite{scalable_NN} in RL, can attain the performance up to $50\%$ sparse network in online training. On the other hand our proposed approach can leverage the fixed batch dataset offline training method and find a $95\%$ pruned sub-architecture. Our proposed method not only shrinks the size of the network weights, with proper hardware and software optimization this will allow faster training and inference \cite{nvidia, dey2019pre}. This reduces the computation cost and allows RL algorithms to use in low-resource, large- data driven real-time applications.

In our initial experiments we tried varying $(i)$ batch size and $(ii)$ number of pruning iterations, but that does not provide any improvement over a single batch pruning loop. Since both methods perform similar to the non-pruned network, we do not conduct further experiments avoid unnecessary compute.

\subsection{Performance of Offline RL algorithm with pruned network}

\subsection{Visualization of the Network After Pruning}
\begin{figure}[hbt!]
\centering
    \subfloat[halfcheetah-expert-v2]{\includegraphics[width=0.32\linewidth]{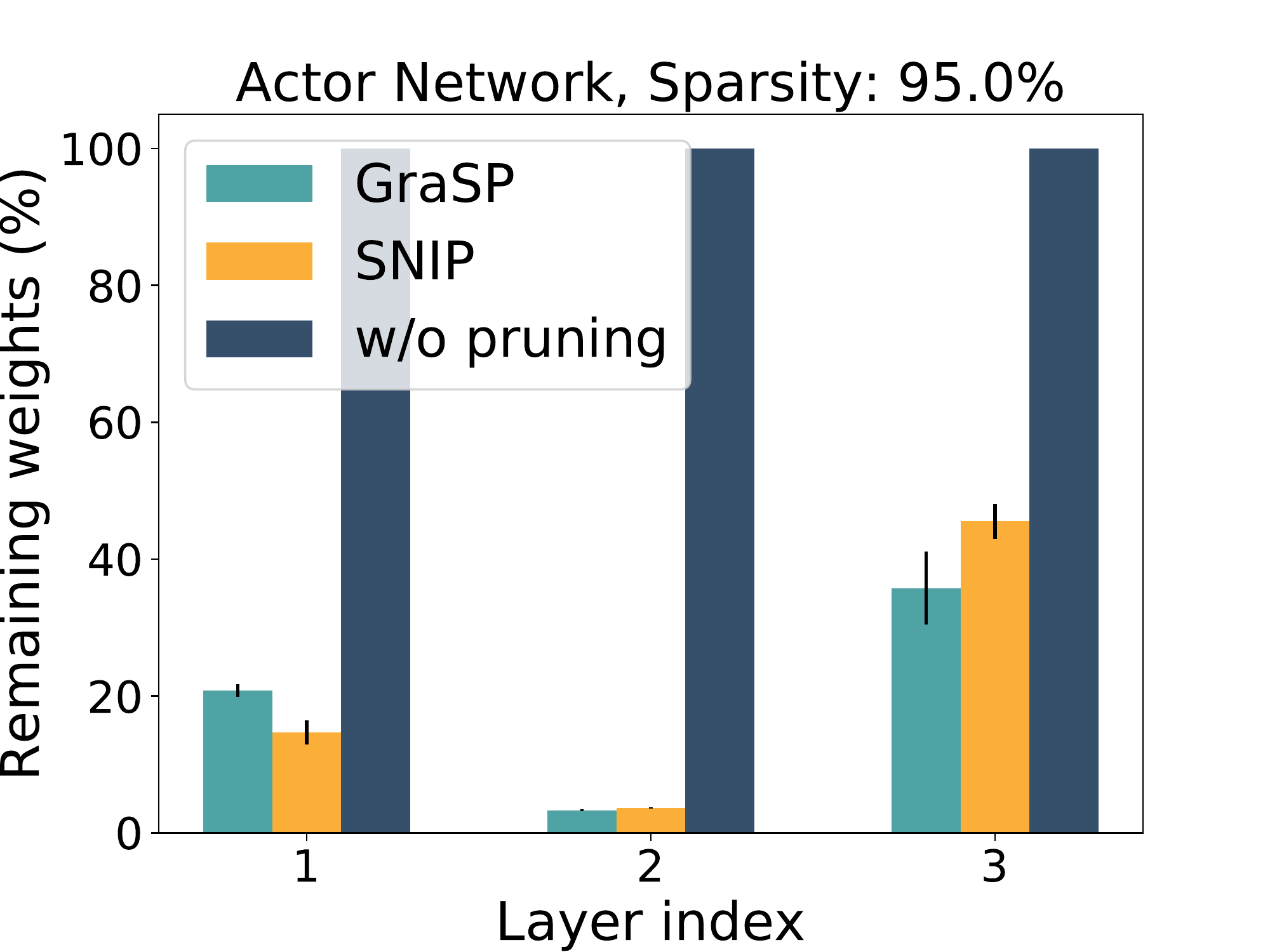}}
   \
    \subfloat[halfcheetah-expert-v2]{\includegraphics[width=0.32\linewidth]{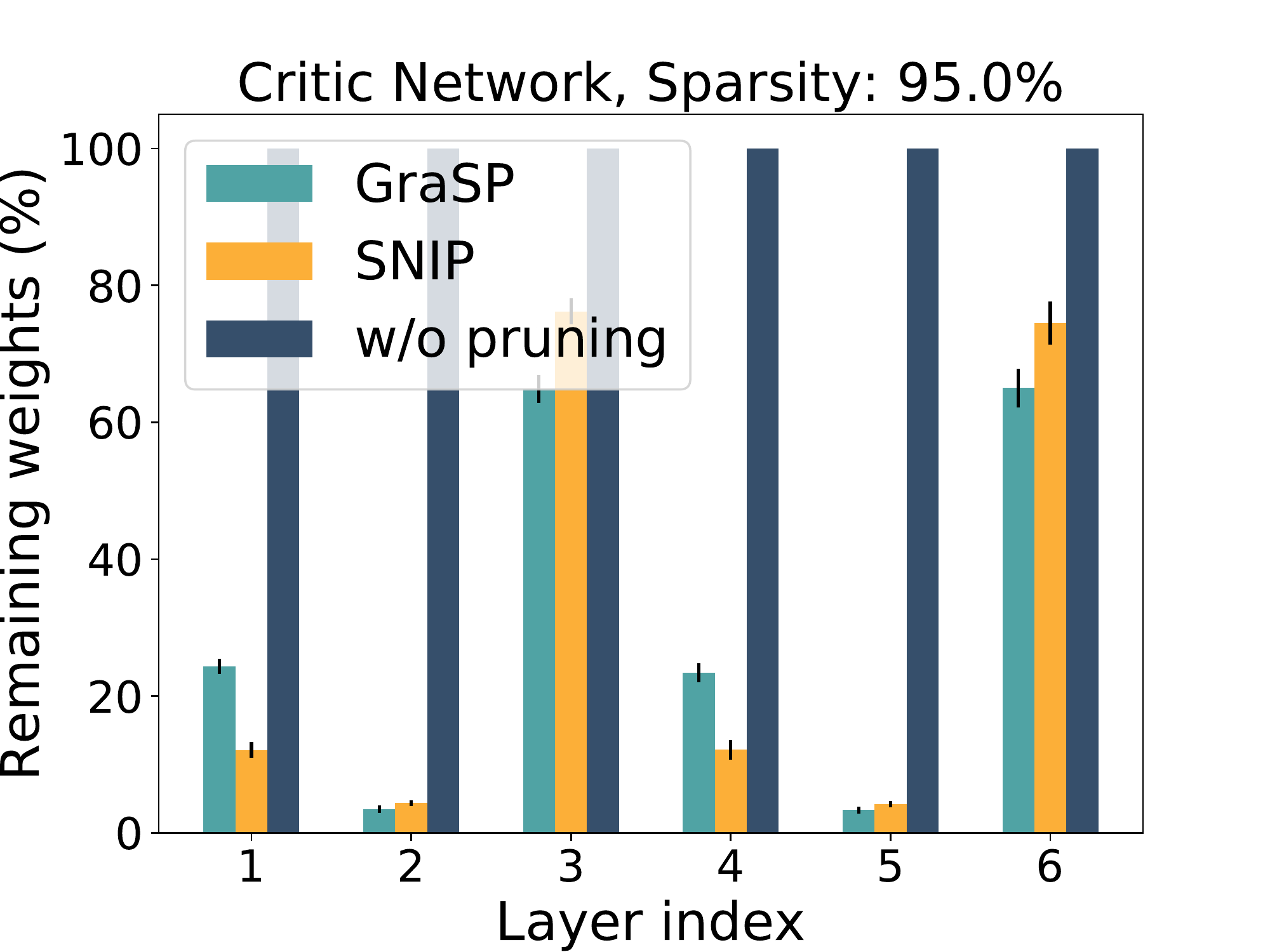}}
   \
    \subfloat[halfcheetah-expert-v2]{\includegraphics[width=0.32\linewidth]{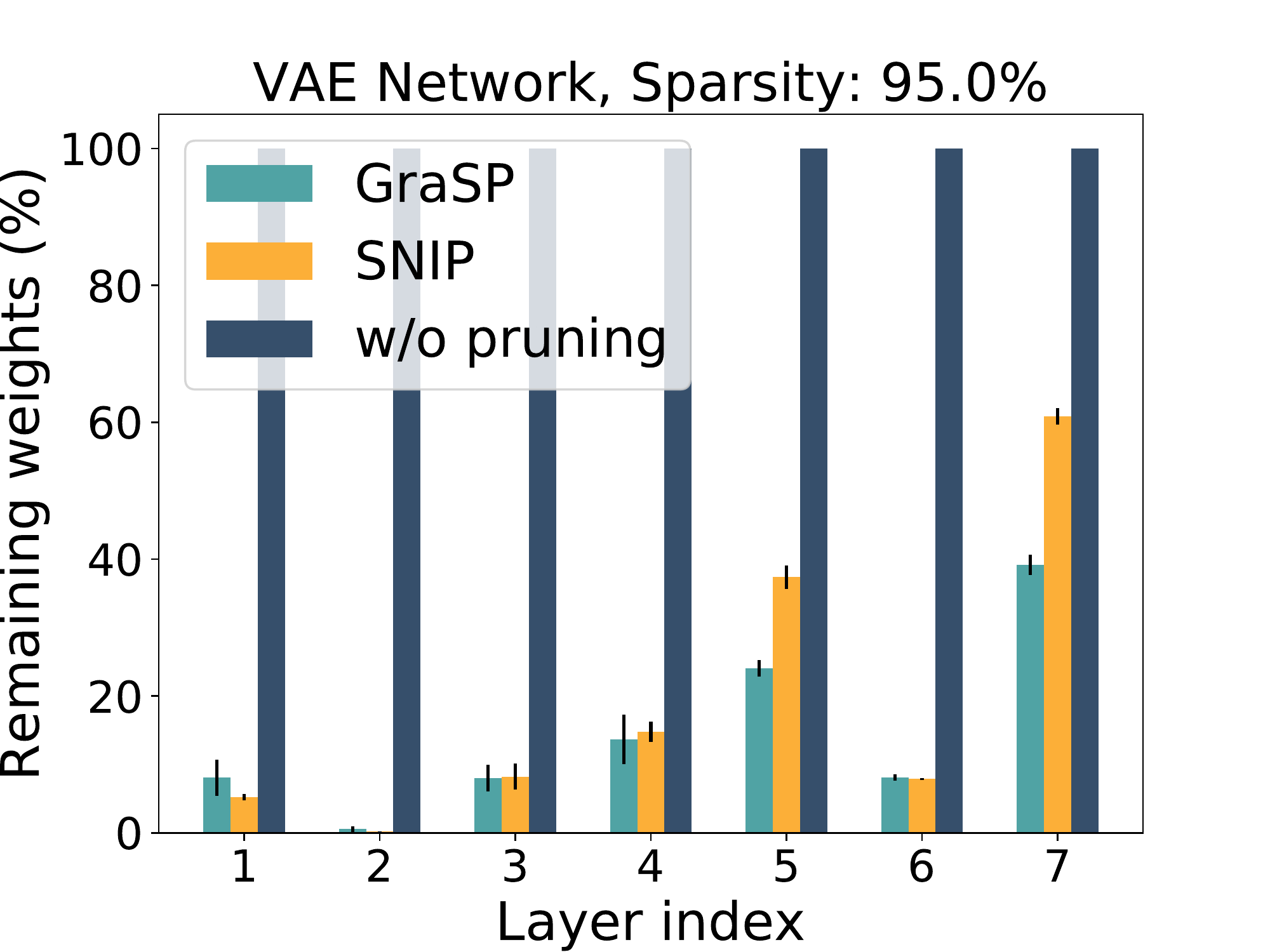}}
    \
    
    \vspace{-0.45cm}
     \subfloat[hopper-expert-v2]{\includegraphics[width=0.32\linewidth]{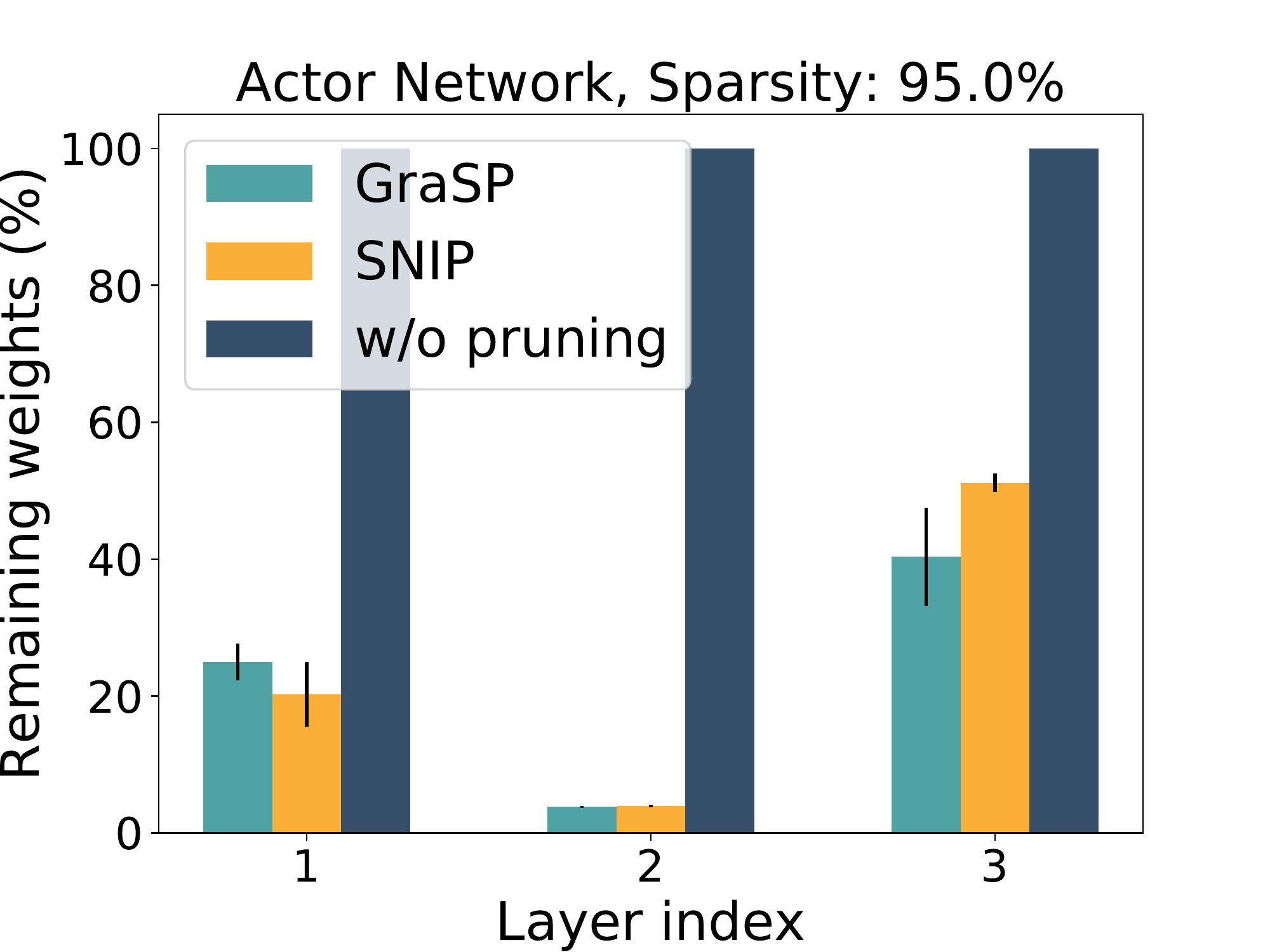}}
   \
    \subfloat[hopper-expert-v2]{\includegraphics[width=0.32\linewidth]{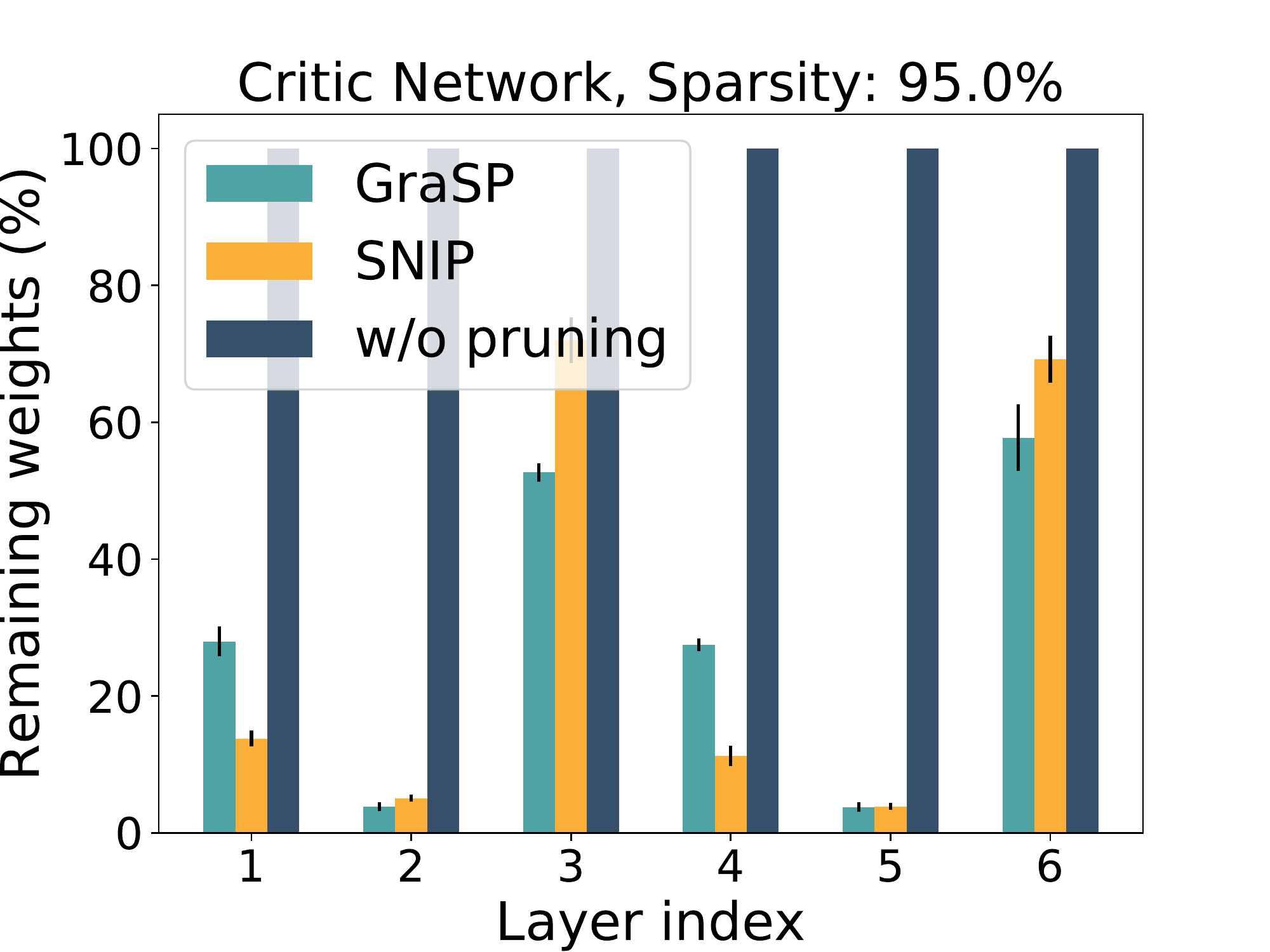}}
   \
    \subfloat[hopper-expert-v2]{\includegraphics[width=0.32\linewidth]{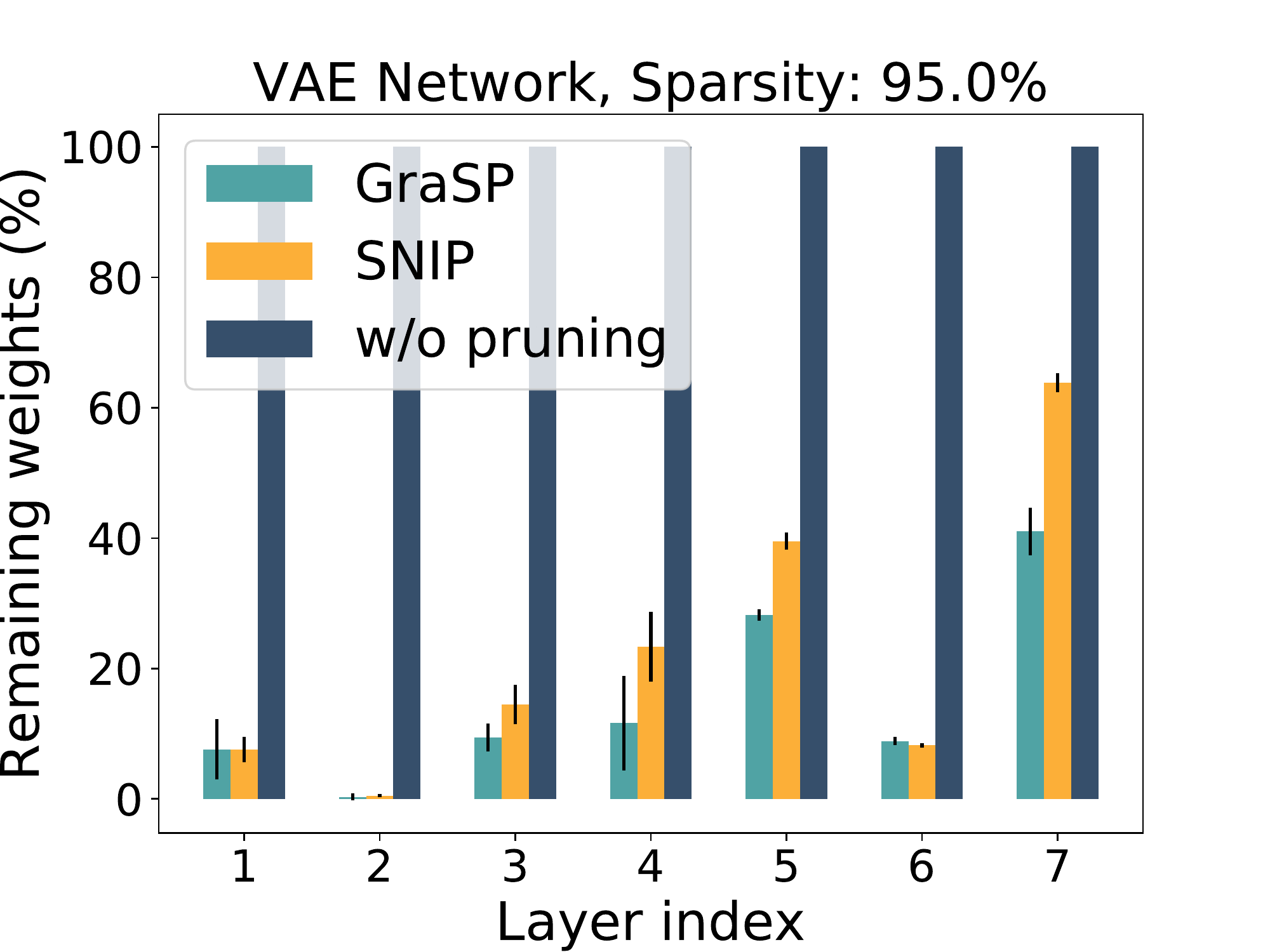}}
    \
    
    \vspace{-0.45cm}
     \subfloat[walker2d-expert-v2]{\includegraphics[width=0.32\linewidth]{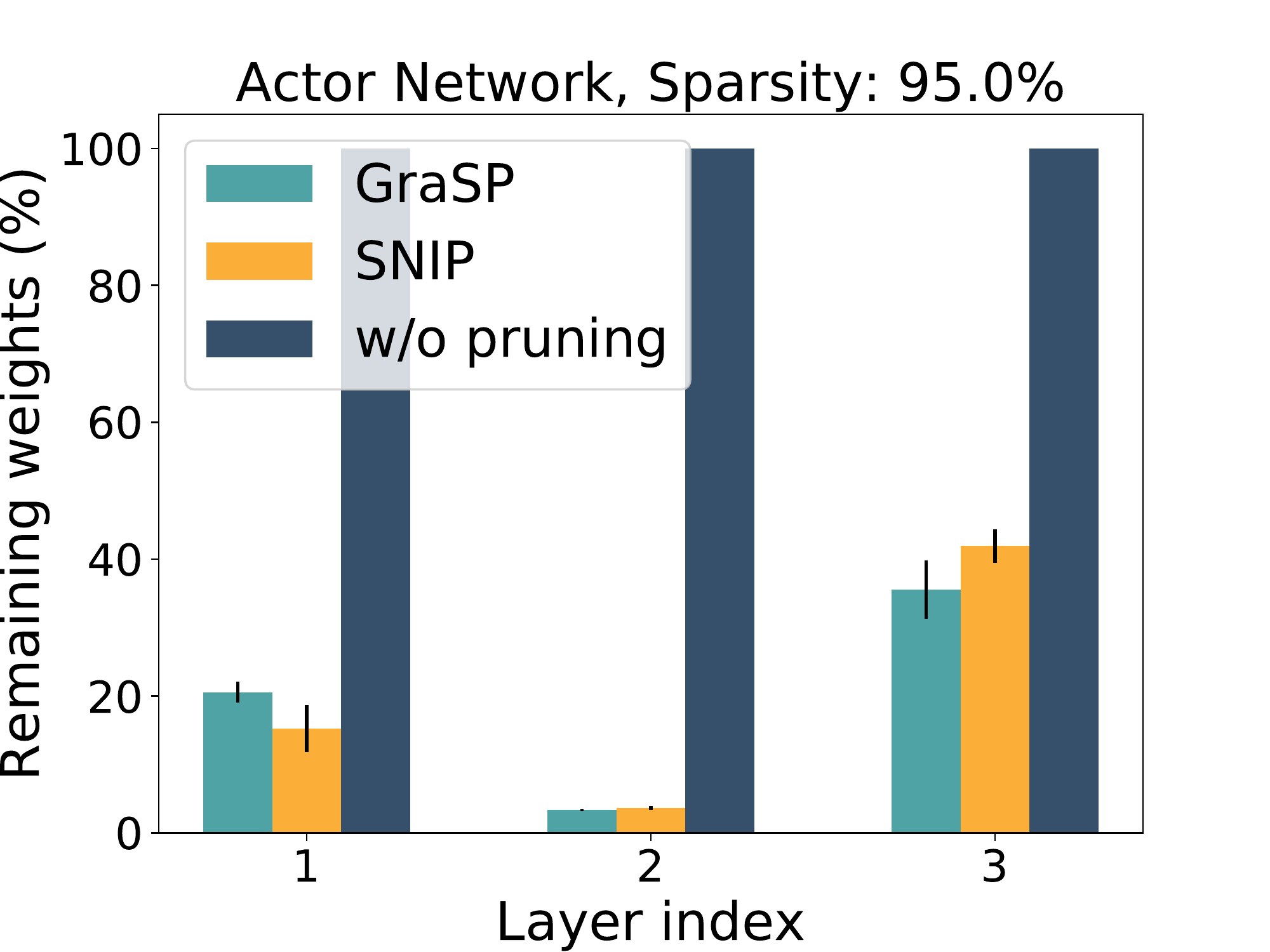}}
   \
    \subfloat[walker2d-expert-v2]{\includegraphics[width=0.32\linewidth]{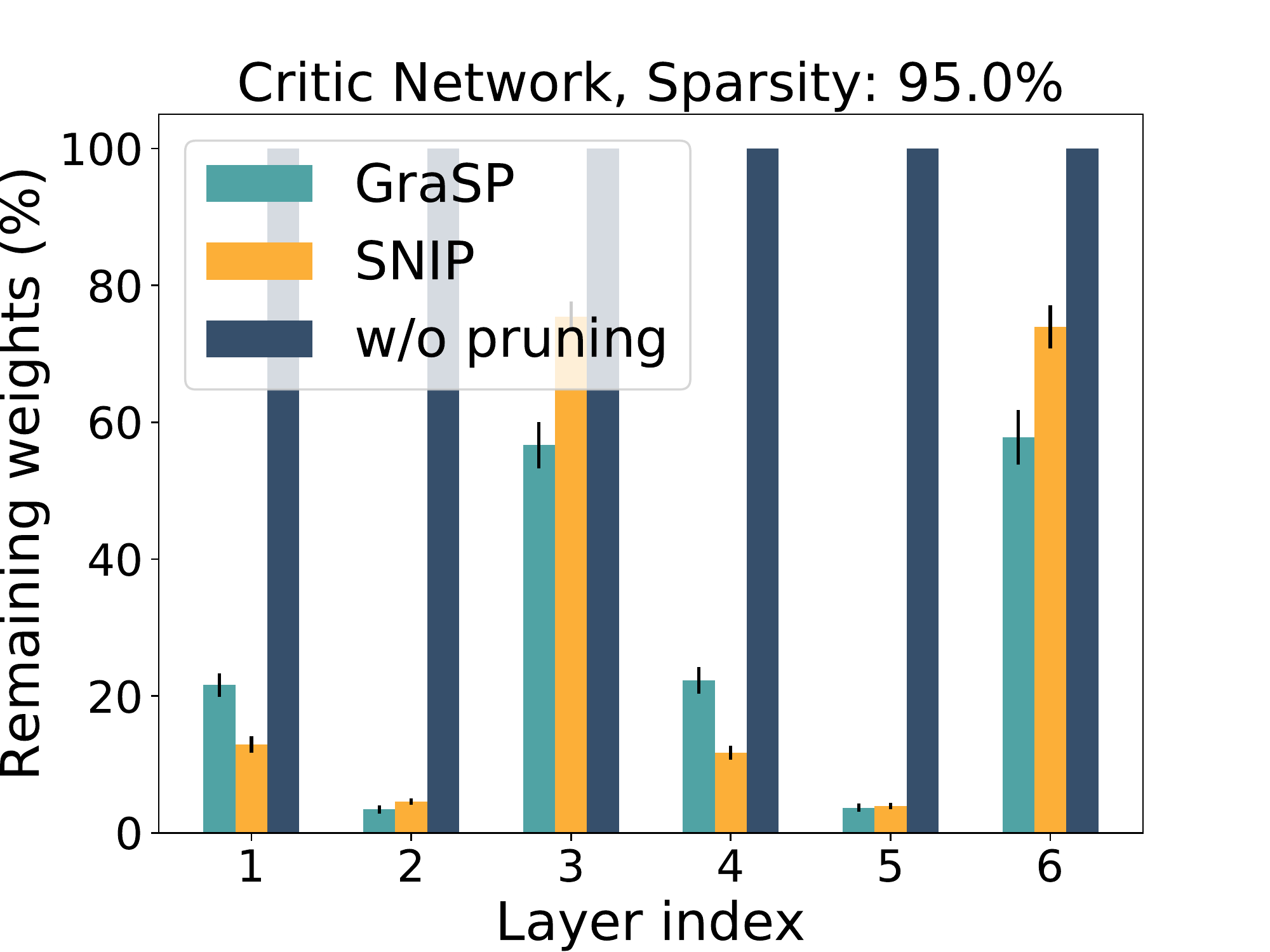}}
   \
    \subfloat[walker2d-expert-v2]{\includegraphics[width=0.32\linewidth]{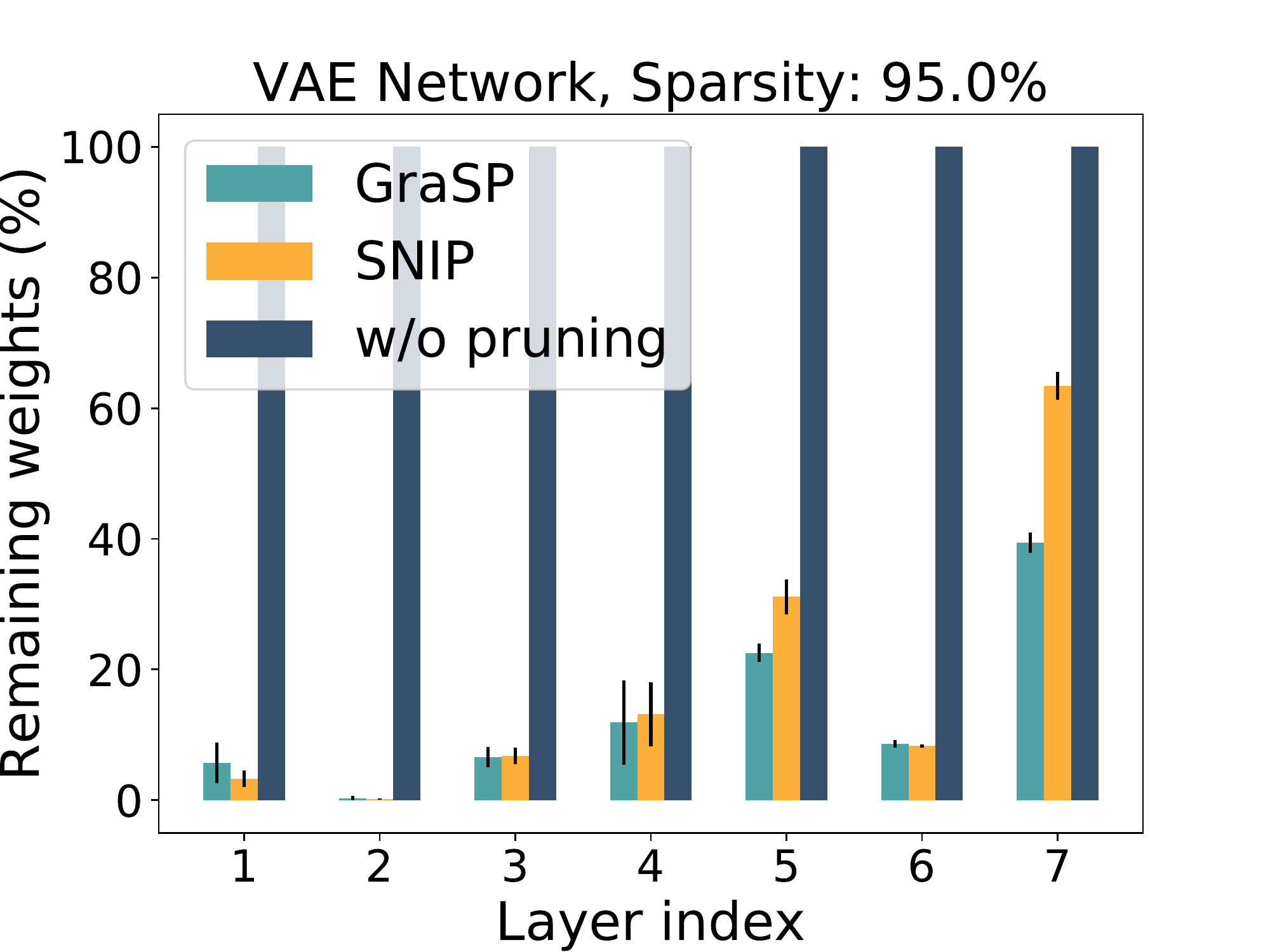}}
\caption{Visualization of the remaining weights per-layer of the neural networks}
\label{Layerwise_sparsity}
\end{figure}

We compare the layer wise sparsity of the pruned network to that of the regular dense network and observe the effect of layer-wise pruning. In figure \ref{Layerwise_sparsity}, we compare the layer-wise remaining weights using different techniques after pruning them with $95\%$ sparsity. We plot the mean number of layer-wise weights and it's standard-deviation for seeds 0-4. For both SNIP and GraSP we find similar pattern of pruning, where it does not prune all the layers uniformly. Both methods preserve more weights for the last layer to preserve gradient flow. But since GraSP's objective focuses on preserving the gradient flow \cite{wang2020picking} they preserve weights at the last layer than SNIP.

\newpage
\subsection{Network Weights Reduction}

\begin{figure}[hbt!]
\centering
   \includegraphics[width=0.32\linewidth]{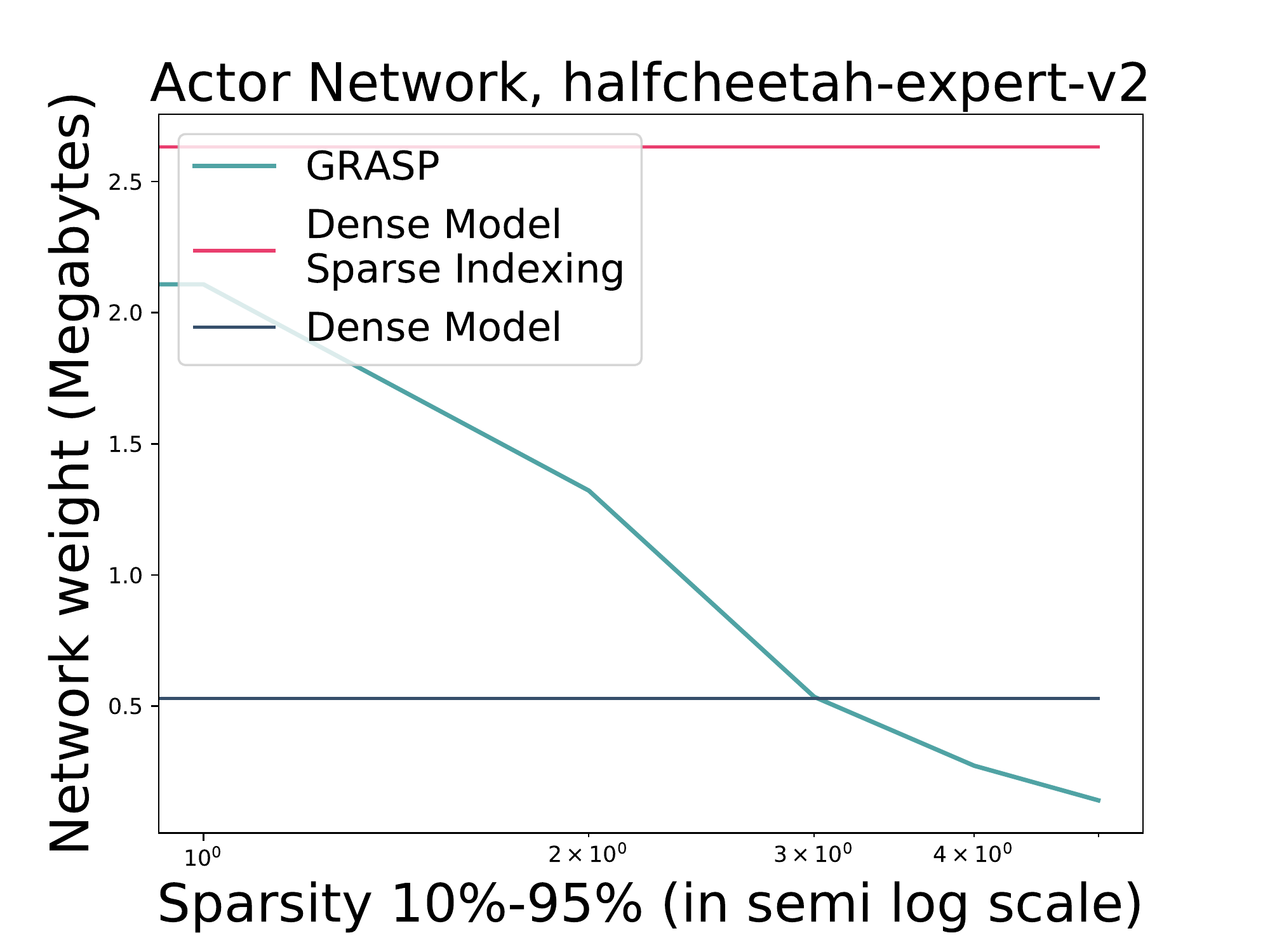}
   \
    \includegraphics[width=0.32\linewidth]{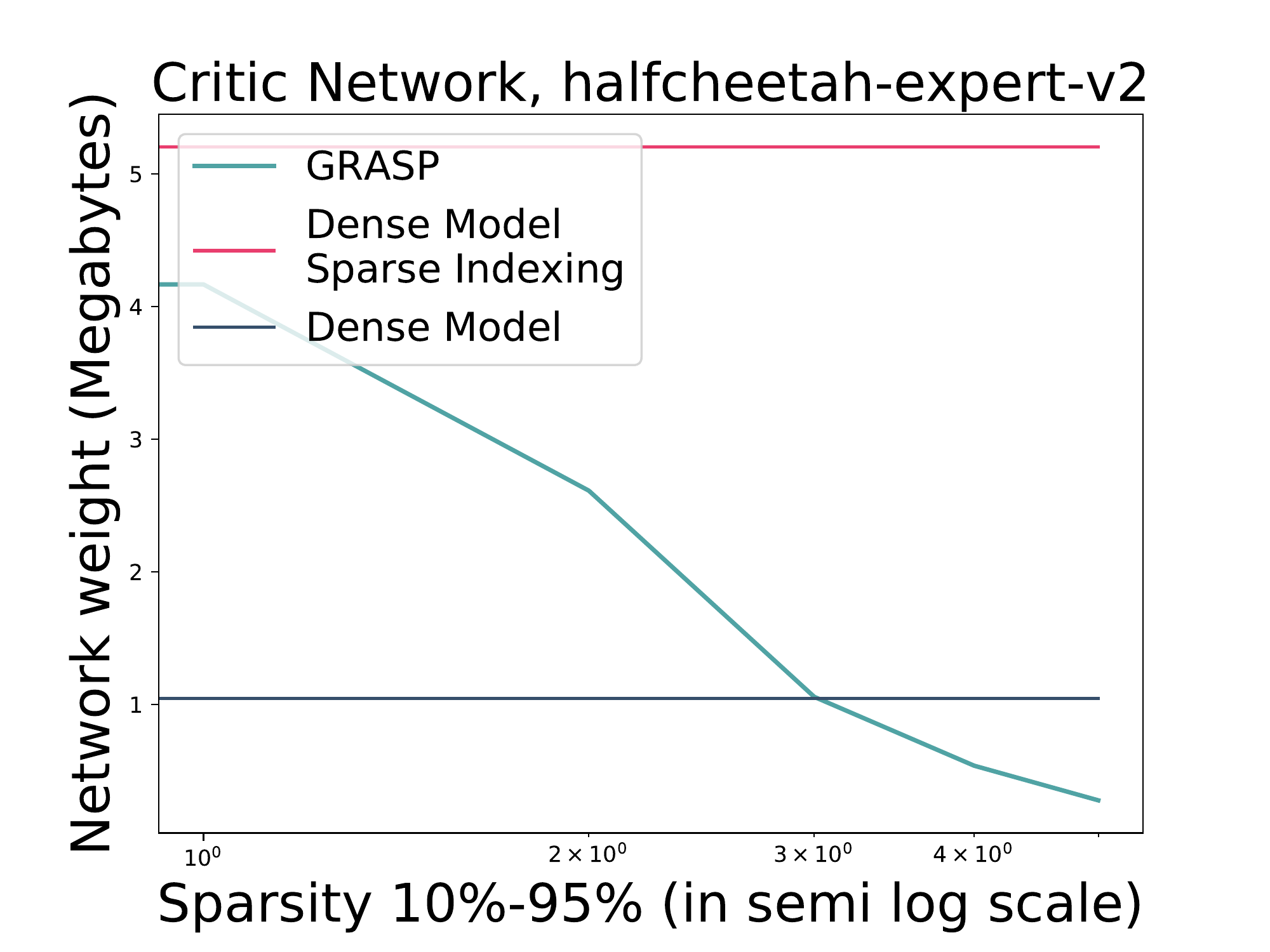}
   \
    \includegraphics[width=0.32\linewidth]{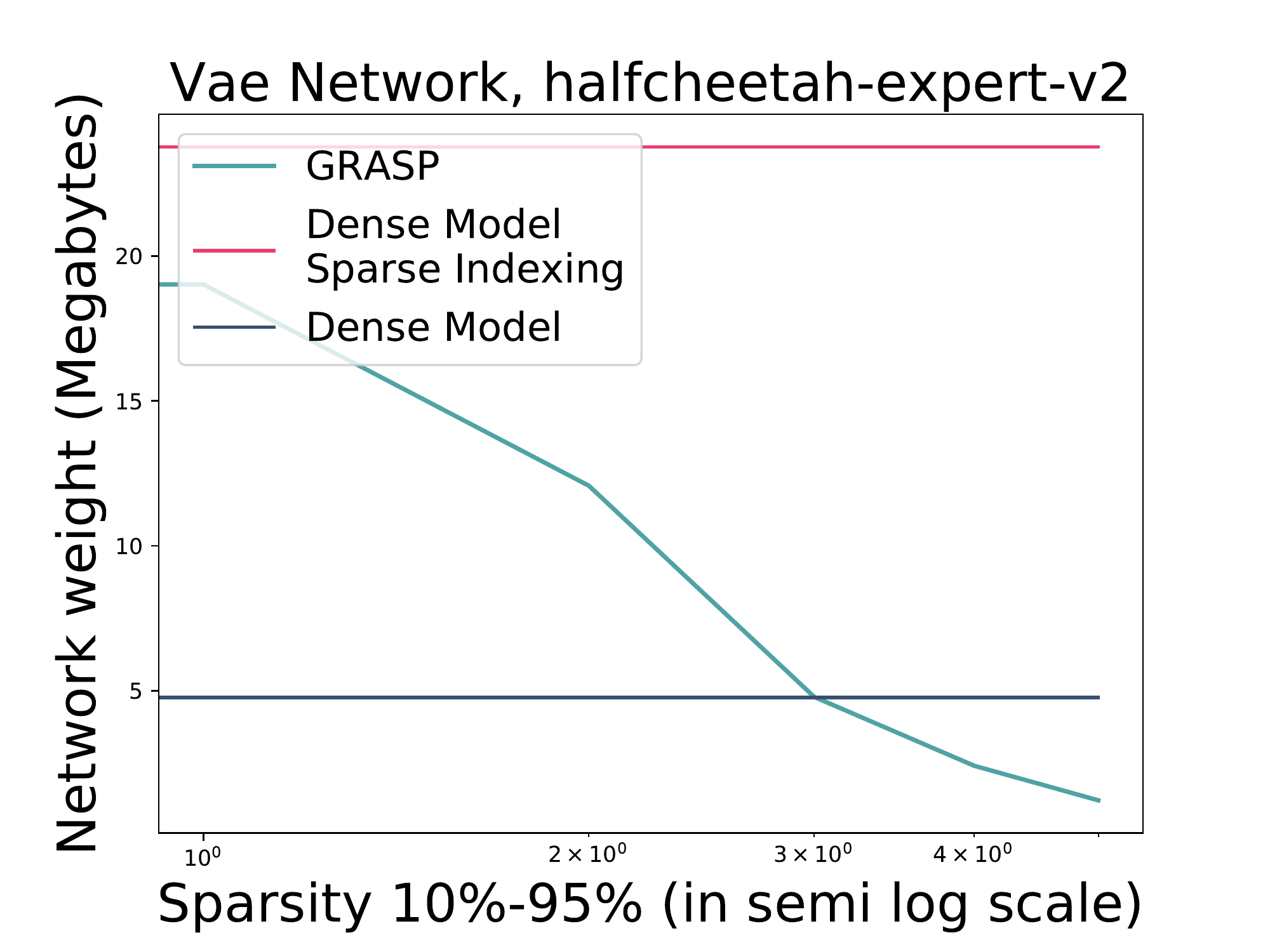}
\caption{Comparison of memory requirement of sparse and dense models}
\label{Layerwise sparsity}
\end{figure}

We use "\emph{torch$.$to$\_$sparse()}" function from PyTorch \cite{pytorch_lib} library to get the sparse matrix, which stores the weights and corresponding index vectors. Since sparse indexing (green line in figure \ref{Layerwise sparsity}) requires additional index vectors, it takes more memory to save regular dense network weights (blue line in figure \ref{Layerwise sparsity}). For $95\%$ sparsity, we are able to reduce the memory size to \emph{$4 \text{x}$} compared to the regular dense networks. With more sophisticated compression mechanisms to save sparse matrices, it will be possible to achieve further reduction in memory requirements. In Table \ref{network-weight-table} we compare the memory (in Megabytes) it takes to save these networks.

\begin{table}
  \caption{Memory Size of the Network Weights in Megabytes (mb)}
  \label{network-weight-table}
  \centering
  \begin{tabular}{llll}
    \toprule
    \cmidrule(r){1-2}
    Method     & Actor & Critic & VAE  \\
    \midrule
    Dense Model                 &  0.5287         &  1.04492 & 4.7621 \\
    Dense Model Sparse Indexing &  2.6318         &  5.2035  & 23.7739  \\
    GraSP (95\% sparse)                      &  \textbf{0.14099}& \textbf{0.2768}  & \textbf{1.2122}  \\
    SNIP  (95\% sparse)                       & \textbf{0.14297}  & \textbf{0.2824} & \textbf{1.2303}  \\
    \bottomrule
  \end{tabular}
\end{table}


\section{Future Work}
We use D4RL \cite{D4RL} dataset for this experiment where expert data were collected from the same data distribution. In real-world application that will not be the case. And one-shot techniques does not guarantee performance under changes in the underlying distribution. 
\section{Conclusion}

Network latency is one of the more crucial aspects of deploying a deep RL into real world application where it needs to process large dataset in real-time (i.e. self-driving car, deploying bot in games, financial data analysis etc.). This also hinders applying RL in low resource devices (i.e. embedded system, edge devices etc.). As a step towards this direction we conduct experiments on pruning techniques in offline RL algorithms. In this paper we show, how simple single-shot pruning plug-in prior to training can reduce the network parameters by $95\%$ while maintaining performance. This sparse model saves $4 \text{x}$ in memory without using any compression mechanism and with proper hardware integration \cite{nvidia, dey2019pre} it promises faster training and higher inference time.

\section{Acknowledgement}

Riyasat Ohib and Sergey Plis were in part supported by R01 DA040487 and RF1 MH121885 from NIH.

\bibliographystyle{unsrt}  

\bibliography{references}

\end{document}